\documentclass[10pt,twocolumn,letterpaper]{article}

\usepackage{iccv}      
\usepackage{times}
\usepackage{epsfig}

\usepackage{graphicx}
\usepackage{amsmath}
\usepackage{amssymb}
\usepackage{booktabs}
\usepackage{appendix}
\usepackage{pifont}
\usepackage{float}
\usepackage{multicol}
\usepackage{pdfpages}

\usepackage{multirow}
\usepackage{bm}
\usepackage[table]{xcolor}

\usepackage[pagebackref,breaklinks,colorlinks]{hyperref}

\usepackage[capitalize]{cleveref}
\crefname{section}{Sec.}{Secs.}
\Crefname{section}{Section}{Sections}
\Crefname{table}{Table}{Tables}
\crefname{table}{Tab.}{Tabs.}

\newcommand{\tabincell}[2]{\begin{tabular} {@{}#1@{}}#2 \end{tabular}}

\iccvfinalcopy 

\def\iccvPaperID{7341}

\ificcvfinal\fi

\begin{document}

\title{Coordinate Transformer: Achieving Single-stage Multi-person \\ Mesh Recovery from Videos}


\author{Haoyuan Li$^{1*}$ \quad
Haoye Dong$^{2}\thanks{Both authors contributed equally to this work as co-first authors.}$ \quad
Hanchao Jia$^3$ \quad
Dong Huang$^2$\quad
Michael C. Kampffmeyer$^4$ \quad \\
Liang Lin$^{5\dag}$ \quad
Xiaodan Liang$^{1,6}$\thanks{Corresponding author.}\\
$^1$Shenzhen campus of Sun Yat-sen University \quad $^2$Carnegie Mellon University \\
$^3$Samsung Research China – Beijing (SRC-B) \quad $^4$UiT The Arctic University of Norway\\
$^5$Sun Yat-sen University \quad $^6$Mohamed bin Zayed University of AI\\
{\tt\small \texttt{lihy285@mail2.sysu.edu.cn, donghaoye@cmu.edu, hanchao.jia@samsung.com, donghuang@cmu.edu}}\\\vspace{-1mm}
{\tt\small\texttt{michael.c.kampffmeyer@uit.no, linliang@ieee.org, xdliang328@gmail.com}}
}



\maketitle
\ificcvfinal\thispagestyle{empty}\fi

\begin{abstract}
Multi-person 3D mesh recovery from videos is a critical first step towards automatic perception of group behavior in virtual reality, physical therapy and beyond. However, existing approaches rely on multi-stage paradigms, where the person detection and tracking stages are performed in a multi-person setting, while temporal dynamics are only modeled for one person at a time. Consequently, their performance is severely limited by the lack of inter-person interactions in the spatial-temporal mesh recovery, as well as by detection and tracking defects. To address these challenges, we propose the \textbf{Coord}inate trans\textbf{Former} (CoordFormer) that directly models multi-person spatial-temporal relations and simultaneously performs multi-mesh recovery in an end-to-end manner. Instead of partitioning the feature map into coarse-scale patch-wise tokens, CoordFormer leverages a novel Coordinate-Aware Attention to preserve pixel-level spatial-temporal coordinate information. Additionally, we propose a simple, yet effective Body Center Attention mechanism to fuse position information. Extensive experiments on the 3DPW dataset demonstrate that CoordFormer significantly improves the state-of-the-art, outperforming the previously best results by 4.2\%, 8.8\% and 4.7\% according to the MPJPE, PAMPJPE, and PVE metrics, respectively, while being 40\% faster than recent video-based approaches. The released code can be found at \href{https://github.com/Li-Hao-yuan/CoordFormer}{https://github.com/Li-Hao-yuan/CoordFormer}

\end{abstract}

\begin{figure}
    \setlength{\belowcaptionskip}{-15pt}
    \centerline{\includegraphics[width=0.48\textwidth]{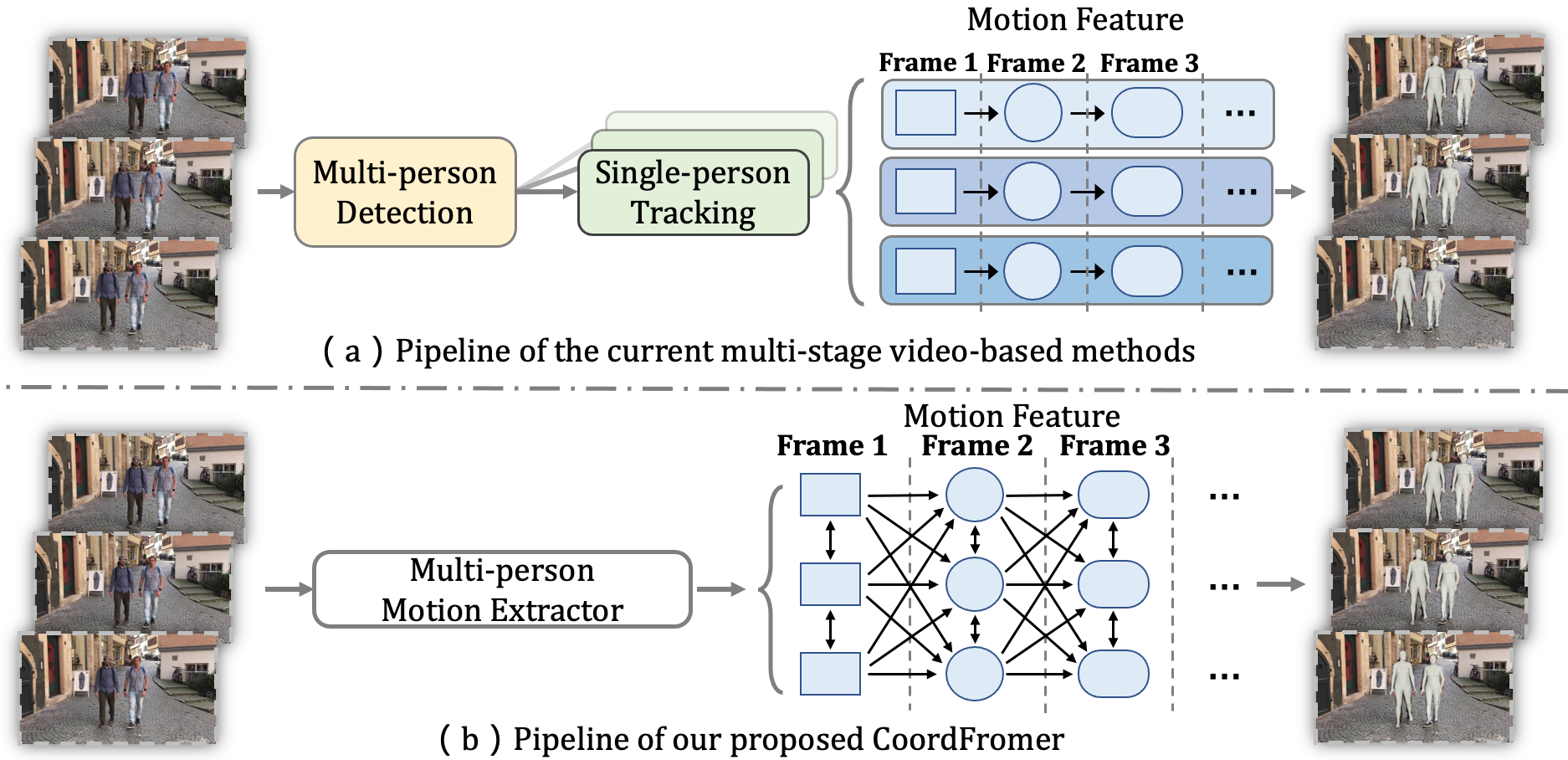}}
    \caption{Comparison of video-based multi-person mesh recovery pipelines. (a) \textbf{Multi-stage} pipelines~\cite{kocabas2020vibe,choi2021beyond,zheng20213d,wei2022capturing,yuan2022glamr}  explicitly generate tracklets and model single-person temporal mesh sequences independently. (b) Our \textbf{single-stage} CoordFormer implicitly matches persons across frames and simultaneously models multi-person mesh sequences in an end-to-end manner.}
    \vspace{-5mm}
    \label{fig:novelty}
\end{figure}

\vspace{-5mm}
\section{Introduction}
\label{sec:intro}
\vspace{-2mm}

Considerable progress has been made on monocular 3D human pose and shape estimation from images~\cite{bogo2016keep,varol2017learning,kanazawa2018end,kolotouros2019learning,xu2019denserac} due to extensive efforts of computer graphics and augmented/virtual reality researchers. However, while frame-wise body mesh detection is feasible, many applications require direct video-based pipelines to avoid spatial-temporal incoherence and missing frame-based detections~\cite{kocabas2020vibe,choi2021beyond,wei2022capturing}.

Existing video-based methods follow a multi-stage design that involves using a 2D person detector and tracker to obtain the image sequences of a single-person for pose and shape estimation~\cite{kanazawa2019learning,kocabas2020vibe,wei2022capturing,zeng2022deciwatch,yuan2022glamr}. More specifically, these methods first detect and crop image patches that contain persons, then track these individuals across frames, and associate each cropped image sequence with a person. The frame-level or sequence-level features are then extracted and used to regress 3D human mesh sequences under spatial and temporal constraints. However, the accuracy of the detection and tracking stage greatly affects the performance of these multi-stage approaches, making them particularly sensitive to false, overlapping, and missing detections.
Moreover, these multi-stage approaches have a considerable computation cost and lack real-time perspectives since the single-person meshes can only be recovered sequence-by-sequence after detection and tracking.

To address the above issues, we introduce CoordFormer, the first single-stage approach for multi-person 3D mesh recovery from videos that can be trained in an end-to-end manner. 
As shown in \cref{fig:novelty}, our method differs from current state-of-the-art approaches~\cite{kocabas2020vibe,choi2021beyond,zheng20213d,wei2022capturing,yuan2022glamr} by being a single-stage pipeline that implicitly performs detection and tracking through the interaction of feature representations, producing multiple mesh sequences simultaneously.

\begin{figure}
    \setlength{\belowcaptionskip}{-15pt}
    \centerline{\includegraphics[width=0.49\textwidth]{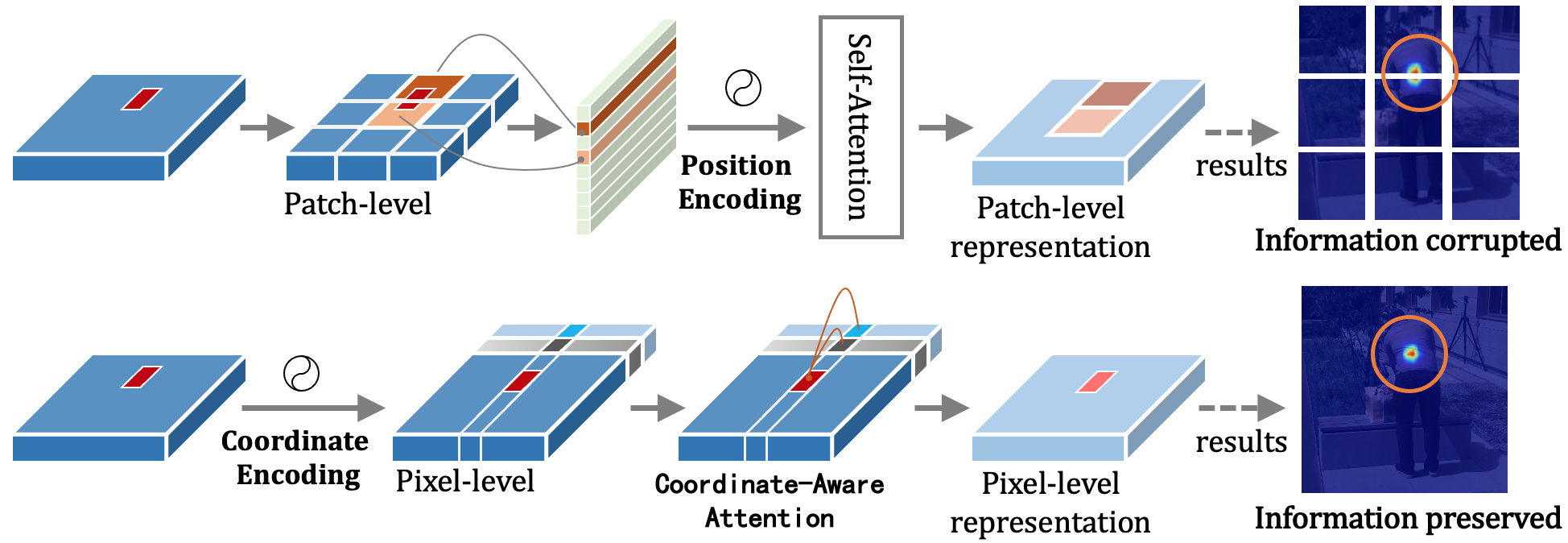}}
    \caption{The motivation of our Coordinate-Aware Attention (CAA) module in CoordFormer. 
    (Top) The standard Transformer based modules (such as ST-Trans~\cite{zheng20213d,liu2022spatial,li2022mhformer}) model patch-level dependency, which results in corruption of pixel-level features.
    (Bottom) CAA encodes pixel-level spatial-temporal coordinates and preserves pixel-level dependencies in features.}
    \vspace{-6mm}
    \label{fig:motivation}
\end{figure}

In particular, CoordFormer leverages a multi-head framework to predict a body center heatmap, which is encoded using our proposed Body Center Attention (BCA). BCA serves as a weak/intermediate person detector that focuses the framework-wide feature representations on potential body centers. Many-to-many temporal-spatial relations among people and across frames are then derived from the BCA-focused features and directly mapped to mesh sequences using our novel Coordinate-Aware Attention (CAA). CAA is integrated into a Spatial-Temporal Transformer (ST-Trans)~\cite{zheng20213d,liu2022spatial,li2022mhformer} to capture non-local context relations at the pixel level. See \cref{fig:motivation} for an illustration of CAAs motivation.
Facilitated by BCA and CAA, CoordFormer advances existing video mesh recovery solutions beyond explicit detection, tracking and sequence modeling. Under various experimental settings on the 3DPW dataset, CoordFormer significantly outperforms the best results of state-of-the-art by 4.2\%, 8.8\% and 4.7\% on MPJPE, PAMPJPE and PVE metrics, respectively. CoordFormer also improves inference speed by 40\% compared to the state-of-the-art video-based approaches~\cite{kocabas2020vibe,wei2022capturing}. Moreover, we demonstrate that enhancing and capturing pixel-level coordinate information significantly benefits the performance under multi-person scenarios. 

The main contributions of this work are as follows:\vspace{-2.5mm}
\begin{itemize}
    \item We propose the first single-stage multi-person video mesh recovery approach, where our BCA mechanism fuses position information and our CAA module enables end-to-end multi-person model training.
    \vspace{-3mm}
    \item We demonstrate that the pixel-level coordinate correspondence is the most critical factor for performance.
    \vspace{-6mm}
    \item Extensive experiments on challenging 3D pose datasets demonstrate that the proposed method achieves significant improvements, outperforming the state-of-the-art methods.
\end{itemize}
\vspace{-1mm}

\vspace{-5mm}

\section{Related Work}
\label{sec:formatting}
\vspace{-2.5mm}

\textbf{Single image-based 3D human pose and shape estimation.} Single image-based methods typically train models to estimate pose, shape, and camera parameters from images, and then output a 3D human mesh using parametric human body models, for example, the SMPL~\cite{loper2015smpl}. Significant progress has been made in this area by leveraging inherent properties of the 3D human, supervising the models using 2D keypoints~\cite{kolotouros2019learning}, semantic segmentation~\cite{jiang2020coherent}, texture consistency~\cite{pavlakos2019texturepose}, interpenetration and depth ~\cite{jiang2020coherent}, body shape~\cite{choutas2022accurate} and IUV maps~\cite{joo2020exemplar}.
However, they primarily use a multi-stage paradigm that is limited by the first stage. BMP~\cite{zhang2021body} improves upon this by proposing a single-stage model that is more robust to occlusions through inter-instance ordinal relation supervision and taking into account body structure. Concurrently, ROMP~\cite{sun2021monocular} adopts a multi-head design which predicts a Body Center heatmap and a Mesh Parameter map. Via parsing the Body Center heatmap and sampling from the Mesh Parameters map, ROMP is able to extract and predict 3D human meshes for multi-person scenarios. BEV~\cite{sun2022putting} extends upon this by further leveraging relative depth information to effectively avoid mesh collision in the single-stage design, as well as age information.
Despite these advances in estimating human pose and shape from single images, these above methods are restricted to single images and poorly capture motion relations of spatial interaction. 

\textbf{Video-based 3D human pose and shape estimation.} 
The existing video-based methods are similarly built based on SMPL and extract SMPL parameters from frames~\cite{sun2019human,kanazawa2019learning,arnab2019exploiting}. However, in these methods, a greater focus is put on modeling temporal consistency and motion coherence. As their image counterparts, video methods follow a two-stage design where people are first detected and features of the bounding-boxes are extracted. In the second stage, tracking is used to capture the motion sequence and refine the pose and shape estimation. More specifically, Sun et. al~\cite{sun2019human} disentangle skeleton features for improving the learning of spatial features and develop a self-attention temporal network for modelling temporal relations. Additionally, they propose an unsupervised adversarial training strategy for guiding the representation learning of motion dynamics in the video. HMMR~\cite{kanazawa2019learning} proposes a temporal encoder that learns to capture 3D human dynamics in a semi-supervised manner, while Arnab \emph{et al.}~\cite{arnab2019exploiting} presents a bundle-adjustment-based algorithm for human mesh optimization and a new dataset consisting of in-the-wild videos. Compared to temporal convolutions and optimization across frames, recurrent structures and attention mechanisms provide superior motion information for mesh regression. VIBE~\cite{kocabas2020vibe} first extracts features from each frame and uses a temporal encoder, i.e. bidirectional gated recurrent units (GRU), to model temporal relations and obtain consistent motion sequences. For more realistic mesh results, the discriminator adopts an attention mechanism to weight the contribution of distinct frames. TCMR~\cite{zhang2021body} proposes the PoseForecast approach composed of GRUs, which integrates and refines static features by fusing pose information from past and future frames to ensure motion consistency. MPS-Net~\cite{wei2022capturing} further extends the non-local concept to capture motion continuity, as well as temporal similarities and dissimilarities. MPS-Net further develops a hierarchical attentive feature integration to refine temporal features observed from past and future frames. However, these methods only optimize the motion of individual people and ignore the spatial interactions among people, which is crucial in multi-person scenarios. CoordFormer, instead, adopts a single-stage design for multi-person mesh recovery, aiming at modeling spatial-temporal relations and constraints across frames.


\begin{figure*}
\setlength{\belowcaptionskip}{-12pt}
\centerline{\includegraphics[width=0.99\textwidth]{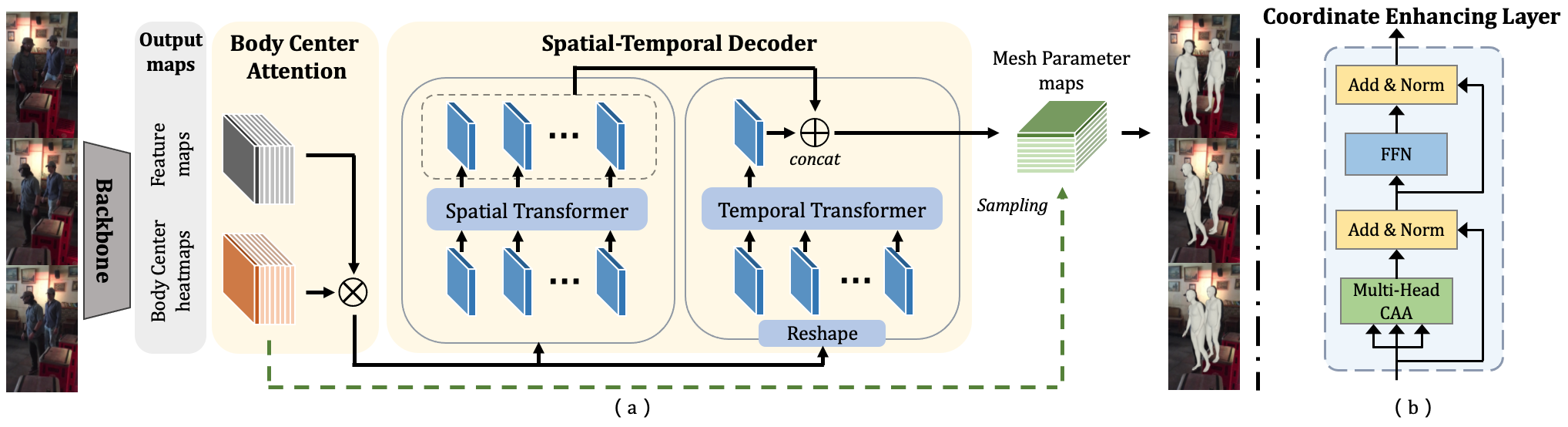}}\vspace{-3mm}
\caption{An overview of the CoordFormer. (a) Given a video sequence, CoordFormer first extracts a Feature map from each image and predicts the Body Center heatmap that reflects the probability of each position being a body center. Then CoordFormer leverages our proposed BCA mechanism and Spatial-Temporal Decoder to predict the pixel-level Mesh Parameter map that contains SMPL and camera parameters. Finally, the Body Center heatmap is parsed and the 3D mesh results are sampled. (b) The Coordinate Enhancing Layer that the Spatial Transformer and the Temporal Transformer of CoordFormer are comprised of. Each layer consist of multi-head CAA operations, a feed-forward network (FFN), Layernorm, and skip connections.
}
\vspace{-5mm}
\label{fig:CoordFormer_overview}
\end{figure*}

\vspace{-3.5mm}
\section{CoordFormer}
\vspace{-2mm}
\textbf{Overview.} We present the CoordFormer framework (see ~\cref{fig:CoordFormer_overview}) to advance multi-person temporal-spatial modelling for video-based 3D mesh recovery. We take inspiration from single-stage image-based approaches for mesh recovery~\cite{sun2021monocular} and leverage a multi-head design that predicts a Body Center heatmap as well as a Mesh parameter map. To further capture the spatial-temporal relations, we introduce two novel modules: (1) the BCA mechanism (\cref{sec:BCA}), which focuses spatial-temporal feature extraction on persons for better performance and faster convergence, and (2) the CAA module (\cref{sec:CEL}) incorporated in a Spatial-Temporal Transformer (\cref{sec:STD}), which preserves pixel-level spatial-temporal coordinate information. CAA avoids the spatial information degradation which usually occurs in the patch-level tokenization of standard vision transformers. 

For completeness and notation consistency we briefly present the Body Center heatmap and the Mesh Parameter map which are predicted by the backbone network. They follow~\cite{sun2021monocular} and are computed for all the $T$ frames in a video.

\textbf{Body Center Heatmap $\textbf{C}_{m}\in \mathbb{R}^{T\times 1\times H\times W}$:} $\textbf{C}_{m}$ (where $H$=$W$=64) represents the likelihood of there being a 2D human body center at a given pixel in the image, where each potential body center is characterised by a Gaussian distribution. Following~\cite{sun2021monocular}, scale information such as body size is encoded in the kernel size of the Gaussian $k$. More specifically, let $d_{bb}$ be the diagonal length of the person bounding box and $W$ be the width of the Body Center heatmap, then $k$ is computed as:

\begin{equation}
	\label{eq:BodyCenterMap}
	k = k_{l}+\frac{\sqrt{2}W}{d_{bb}}^{2}k_{r},
\end{equation}
where $k_{l}$ is the minimum kernel size, $k_{r}$ is the range of $k$.

Note, for in-the-wild images of multiple people, $\textbf{C}_{m}$ not only contains the scale information of every potential target, but also contains strong location information that can be leveraged to reduce redundancy and focus features. This is further explored in Sec.~\ref{sec:BCA}.

\textbf{Mesh Parameter map $\textbf{P}_{m}\in \mathbb{R}^{T\times 145\times H\times W}$:} $\textbf{P}_{m}$ (where $H$=$W$=64) contains the camera parameters $\textbf{A}_{m} \in \mathbb{R}^{T\times 3\times H\times W}$ and SMPL parameters $\textbf{S}_{m}  \in \mathbb{R}^{T\times 142\times H\times W}$. 
	\begin{itemize}
	\item In terms of camera parameters, $\textbf{A}_{m}=(\xi,t_{x},t_{y})$ describes the 2D scale and translation information for every person in each frame, such that the 2D projection $\hat{\emph{\textbf{J}}}$ of the 3D body joints $\emph{\textbf{J}}$ can be obtained as $\hat{\emph{\textbf{J}}_{x}}$ = $\xi \emph{\textbf{J}}_{x}+t_{x}$, $\hat{\emph{\textbf{J}}_{y}}$ = $\xi \emph{\textbf{J}}_{y}+t_{y}$.

	\item The SMPL parameters, $\textbf{S}_{m}$, describe the 3D pose $\bm{\theta}$ and shape $\bm{\beta}$ of the body mesh at each 2D position. For every potential person, $\bm{\theta} \in \mathbb{R}^{6\times 22}$ describes the 3D rotations in the 6D representation~\cite{zhou2019continuity} of each body joint apart from the hands, and $\bm{\beta} \in \mathbb{R}^{10}$ are the shape parameters. Combining $\bm{\theta}$ with $\bm{\beta}$, SMPL establishes an efficient mapping to a human 3D Mesh $\emph{\textbf{M}} \in \mathbb{R}^{6890\times 3}$.
	\end{itemize}

\subsection{BCA: Body Center Attention} 
\label{sec:BCA}
	The Body Center Attention mechanism is at the core of CoordFormer. It aims to fuse position information and acts as a learnable feature indexer by leveraging the representation pattern of the body center heatmap $\textbf{C}_{m}$. Each pixel in $\textbf{C}_{m}$ represents a potential person and learning relations at this pixel-level through Multi-Head Self-Attention (MHSA) would result in redundant calculations as most pixels do not contain people. Instead, we leverage the fact that $\textbf{C}_{m}$ contains effective position information which can be used as a natural additional attention map for locating people in the corresponding frame. We thus use the \textbf{Body Center} heatmap as the \textbf{Attention} map, i.e. Body Center Attention, to focus and extract features of all persons.

    Specifically, given an input video sequence $\textbf{V} = \{\textbf{I}_{t}\}^{T}_{t=1}$ with $T$ frames, we first use the backbone to extract the feature map $\textbf{F}_{m} \in \mathbb{R}^{T\times H\times W\times C}$. To enhance the perception of the coordinate system, we extend $\textbf{F}_{m}$ with coordinate channels~\cite{liu2018intriguing} resulting in $\textbf{F}_{coord}$ and predict $\textbf{C}_{m}$ from it. Finally, we compute the focused features as the Hadamard product between $\textbf{C}_{m}$ and $\textbf{F}_{m}$. Note, here we leverage $\textbf{F}_{m}$ instead of $\textbf{F}_{coord}$, to avoid altering the coordinate features of $\textbf{F}_{coord}$.
	
	Let $\textbf{F}_{m}^{t} \in \mathbb{R}^{H\times W\times C}$, $\textbf{F}_{coord}^{t} \in \mathbb{R}^{H\times W\times (C+2)}$ and $\textbf{C}_{m}^{t} \in \mathbb{R}^{H\times W\times 1}$ be the feature map, coordinate feature map and Body Center heatmap of the $t^{th}$ frame, respectively. The focused feature map of the $t^{th}$ frame $\textbf{F}_{focus}^{t} \in \mathbb{R}^{H\times W\times C}$ can then be computed as follows,
\begin{equation}
	\label{eq:BCA-AddCoordinateChannel}
	\textbf{F}_{coord}^{t} = ACC(\ \textbf{F}_{m}^{t}\ ),
\end{equation}
\begin{equation}
	\label{eq:BCA-Centermap}
	\textbf{C}_{m}^{t} = Conv_{c}(\ \textbf{F}_{coord}^{t}\ ),
\end{equation}
\begin{equation}
	\label{eq:BCA-Focus}
	\textbf{F}_{focus,c}^{t} = \odot(\ \textbf{C}_{m}^{t}, \textbf{F}_{m,c}^{t}\ ),
\end{equation}
where $ACC(\cdot)$ indicates adding the coordinate channels, $\odot(\cdot,\cdot)$ indicates the Hadamard product, $Conv_{c}(\cdot)$ is the head convolution layers to obtain the Body Center heatmap, and $\textbf{F}_{focus,c}^{t}$ and $\textbf{F}_{m,c}^{t}$ indicate the $c^{th}$ channel of $\textbf{F}_{focus}^{t}$ and $\textbf{F}_{m}^{t}$, respectively.
	
	As obtaining $\textbf{C}_{m}$ is arguably the simplest learning task in the multi-head framework, it represents a reliable source to obtain the focused features $\textbf{F}_{focus}$ and facilitates the effectiveness of BCA.
%
\begin{figure*}
    \setlength{\belowcaptionskip}{-12pt}
	\centerline{\includegraphics[width=0.99\textwidth]{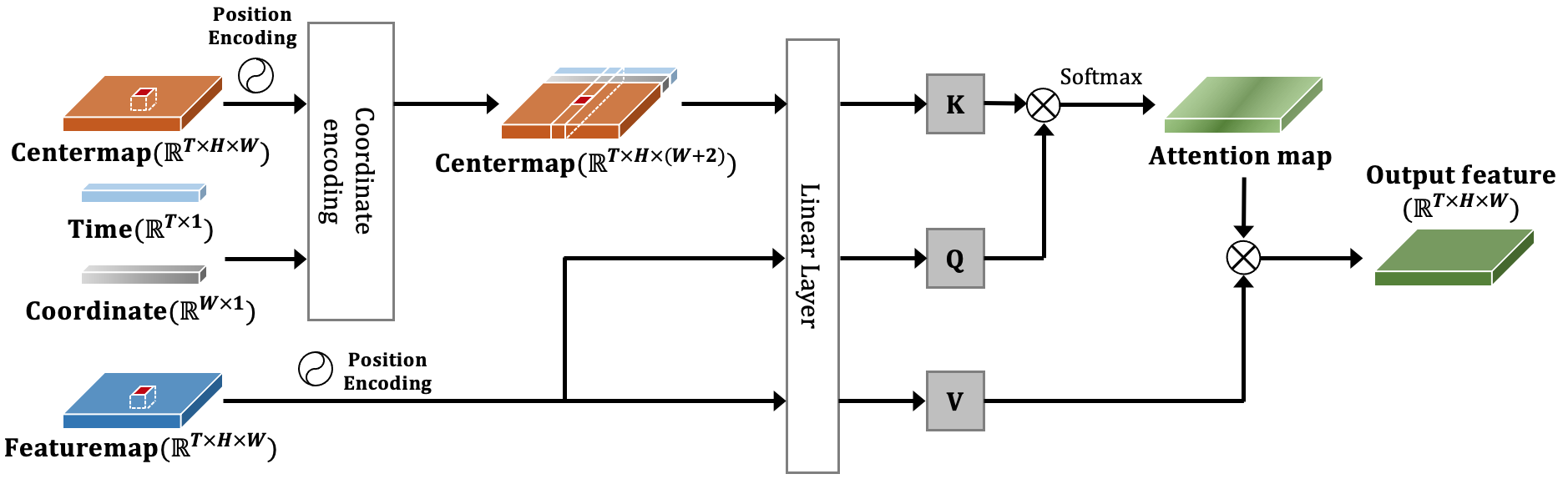}}\vspace{-2mm}
	\caption{Network structure of our CAA module. Given the Centermap and Featuremap as input, the precise coordinate information is encoded in the Centermap by coordinate-encoding and the rough position information is encoded in the Featuremap by Position-Encoding. Then, $K$, $Q$ and $V$ are computed for scaled dot-product attention. With powerful position information as key, CAA can capture high-quality spatial-temporal correspondence among multiple persons.}
    \vspace{-3mm}
	\label{fig:CAA}
\end{figure*}

\subsection{CEL : Coordinate Enhancing Layer}   
\label{sec:CEL}
\vspace{-1mm}
    After establishing the existence and location of the people in the video, the motion sequence features must be used to determine their temporal relationships. Moreover, in multi-person scenarios, it is imperative to understand the spatial-temporal interactions to facilitate accurate mesh recovery. The spatial-temporal constraints between all known entities must therefore be modeled effectively.
	
	Inspired by the progress on Spatial-Temporal Transformers (ST-Trans) with joint coordinates as input~\cite{liu2022spatial,zheng20213d,zhu2022motionbert}, we adopt a powerful ST-Tran as the base model for our Spatial-Temporal Decoder. However, directly applying a ST-Tran on $\text{F}_{focus}$ does not produce the desired results. This is because the patch-level position information captured from the Position-Encoding~\cite{vaswani2017attention} is not enough to regress the precise joint coordinates required for our single-stage design.
    Moreover, as illustrated in~\cref{fig:motivation}, vision transformers~\cite{dosovitskiy2020image} that split features into patches and extract tokens from them, can lead to a degradation in the pixel-level information, especially for $\textbf{C}_{m}$. Empirical evidence for this is provided in the supplementary material.
	
	To add precise coordinate information across frames and maintain the pixel-level representation of $\textbf{C}_{m}$ and $\textbf{P}_{m}$, we introduce the CAA module that expands the self-attention operation of the Transformer Encoder~\cite{vaswani2017attention}. Unlike Position-Encoding~\cite{vaswani2017attention}, which provides only rough location information at the patch-level, the CAA module captures the the coordinate relationships between $( t, x, y )$ of each pixel.
	As depicted in \cref{fig:CAA}, we extend $\textbf{C}_{m}$ with both time and axis coordinates, enabling us to leverage $\textbf{C}_{m}$ for detection while also leveraging the coordinate features to capture relations.
	
	Specifically, we set Pixel coordinate $\textbf{PC} \in \mathbb{R}^{W\times 1}=[1,2,3...,W]$ and Time coordinate $\textbf{TC} \in \mathbb{R}^{T\times 1}=[1,2,3...,T]$ and repeat them to $\textbf{PC}_{r} \in \mathbb{R}^{T\times W\times 1}$ and Time coordinate $\textbf{TC}_{r} \in \mathbb{R}^{T\times H\times 1}$. The input feature $\textbf{F}_{input}^{1} \in \mathbb{R}^{T\times H\times (W+2)}$ of the $1^{th}$ CEL is then the concatenation of $\textbf{F}_{focus}$, $\textbf{PC}_{r}$, and $\textbf{TC}_{r}$.
	
	After adding coordinate information, three linear projections, $f_{Q}, f_{K}, f_{V}$, are applied to transfer $\textbf{C}_{m}$ and $\textbf{F}_{input}^{l}$ into three matrices of equal size, namely the query $\textbf{Q}$, the key $\textbf{K}$, and the value $\textbf{V}$, respectively. The CAA operation is then calculated by: 
\begin{equation}
	\label{eq:CA-QKV}
	\left\{
	\begin{array}{lll}
	\textbf{Q} = f_{Q}(\ \textbf{F}_{input}\ ),\\
	\\
	\textbf{K} = f_{K}(\ \textbf{C}_{m}\ ),\\
	\\
	\textbf{V} = f_{V}(\ \textbf{F}_{input}\ ),
	\end{array}
	\right.
\end{equation}
\begin{equation}
	\label{eq:CA-Attentioon}
	CAA(\ \textbf{C}_{m}, \textbf{F}_{input}\ ) = Softmax(\frac{\textbf{Q}\textbf{K}^{T}}{\sqrt{D}})V.
\end{equation}
    
    As shown in \cref{fig:CoordFormer_overview}, in our proposed CEL, $H$ heads of CAA are applied to $\textbf{C}_{m}$ and $\textbf{F}_{input}^{l}$. Therefore, the output of the $l^{th}$ CEL, $\textbf{F}_{output}^{l} \in \mathbb{R}^{T\times H\times W}$, can be compute as
\begin{equation}
	\label{eq:CEL:1}
	\textbf{F}' = LN( \textbf{F}_{input}^{l} + CAA(\textbf{F}_{input}^{l}) ),
\end{equation}
\begin{equation}
	\label{eq:CEL:2}
	\textbf{F}_{output}^{l} = LN( \textbf{F}' + FFN(\textbf{F}') ),
\end{equation}
	where $LN$ indicates the Layer Normalization~\cite{ba2016layer} and $FFN$ indicates a feed-forward network. The output of layer $l$, $\textbf{F}_{output}^{l}$, is then provided as input to the next layer, i.e. becomes $\textbf{F}_{input}^{(l+1)}$. Through multiple CELs, our Spatial-Temporal Transformers receives sufficient global location information for implicit feature matching.

\vspace{-1.5mm}
\subsection{Spatial-Temporal Decoder}
\vspace{-2.5mm}
\label{section:STD} 
\label{sec:STD}
	Building on the coordinate-awareness induced by CAA, we leverage the Spatial-Temporal Transformers to learn the spatial and temporal constraints, respectively. As shown in \cref{fig:CoordFormer_overview}, the Spatial-Temporal Decoder forms a residual structure and first establishes spatial feature relationships, before modelling the temporal connections.
	
	\textbf{Spatial Transformer Module.}
	Since the representation patterns of the Body Center Heatmap bring spatial information to the feature due to its similarity around the body center, we first use the Spatial Transformer to extract the corresponding spatial information. Given the input $\textbf{F}_{input}$, the Spatial Transformer performs a CAA operation on each frame, where $Q, K, V$ are $\in \mathbb{R}^{k_{t}\times E_{k_{t}}}$, and where $k$ and $E_{k_{t}}$ indicate the number of tokens and the length of the token embedding, respectively.
	
	\textbf{Temporal Transformer Module. }After building the spatial relationships, the Temporal Transformer is used to ensure consistency in the temporal relationships. Given the input $\textbf{F}_{input}$, the Temporal Transformer performs a CAA operation on all frames, where $Q, K, V$ are $\in \mathbb{R}^{(T \cdot k_{t})\times E_{k_{t}}}$.
	
	\textbf{Coordinate Information Fusion.} Since the Coordinate encoding only adds spatial coordinates for one dimension at a time in the 2D image, we observe improvements by transposing $\textbf{C}_{m}$ at alternating layers, thereby infusing coordinate information along both spatial dimensions.
	
    More specifically, each Transformer has $2L$ CELs. At every $L_{2N}^{th}$ layer where $N=[1,2,3...,L]$, $\textbf{C}_{m}$ will be transposed to $\textbf{C}_{mt} \in \mathbb{R}^{T\times W \times H}$ to add precise coordinate information, resulting in
\begin{equation}
	\label{eq:Cm-transposition}
	\begin{aligned}
	\left\{
	\begin{array}{ll}
	\textbf{F}^{l+1}_{output} = CEL(\textbf{C}_{mt}, \textbf{F}^{l}_{input}), &l= 2N \\
	\\
	\textbf{F}^{l+1}_{output} = CEL(\textbf{C}_{m}, \textbf{F}^{l}_{input}), &otherwise.\\
	\end{array}
	\right.
	\end{aligned}
\end{equation}
	
	Through multiple layers of CEL, the Transformer learns the correspondence along both dimensions.

\subsection{Loss Functions}
\vspace{-1mm}

    The loss function of CoordFormer consists of a set of temporal and spatial loss functions that ensure temporal consistency and spatial accuracy, respectively. 
    
	\textbf{Temporal loss} $L_{tem}$. We add $L_{tem}$ to maintain the similarity of adjacent frames via
\begin{equation}
	\label{eq:loss-temporal_loss}
	L_{tem}  = w_{accel}L_{accel} + w_{aj3d}L_{aj3d} + w_{sm}L_{sm},
\end{equation}
	where $L_{accel}$ and $L_{aj3d}$ are the Accel error~\cite{kocabas2020vibe} and the $L_{2}$ loss of the 3D joints offsets, respectively, and $L_{sm}$ is a regular $L_{1}$ loss between consecutive frames, preventing mutation of $\textbf{C}_{m}$ and $\textbf{F}_{m}$. For each loss item, $w_{(\cdot)}$ indicates the corresponding weight.

	\textbf{Spatial losses} $L_{spa}$. For spatial accuracy, we follow the previous methods~\cite{kanazawa2018end,sun2021monocular} to add loss functions on SMPL parameters, 3D body joints, 2D body joints and Center Body heatmap. Specifically, $L_{cm}$ is the focal loss~\cite{sun2021monocular} of the Center Body heatmap. $L_{\theta}$ and $L_{\beta}$ are $L_{2}$ loss of SMPL pose $\overrightarrow{\theta}$ and shape $\overrightarrow{\beta}$ parameters respectively. $L_{prior}$ is the Mixture Gaussian prior loss~\cite{bogo2016keep,loper2015smpl} of the SMPL parameters for supervision of prior knowledge. To supervise the accuracy of the joint prediction, $L_{j3d}$ and $L_{pj2d}$ are added. $L_{j3d}$ consist of $L_{mpj}$ and $L_{pmpj}$, where $L_{mpj}$ is the $L_{2}$ loss of predicted 3D joints $\overrightarrow{J\ \ }$ and $L_{pmpj}$ is the $L_{2}$ loss of the predicted 3D joints after Procrustes alignment with the ground truth~\cite{sun2021monocular,sun2019human}. $L_{pj2d}$ is the $L_{2}$ loss of the 2D projection of the 3D joints $\overrightarrow{J\ \ }$. For each loss item, $w_{(\cdot)}$ indicates the corresponding weight and $L_{spa}$ can be computed as,
\begin{equation}
	\label{eq:loss-spatial-loss}
	\begin{aligned}
	L_{spa}  = &w_{cm}L_{cm} + w_{pose}L_{pose} + w_{shape}L_{shape} \\
	+ &w_{prior}L_{prior} + w_{j3d}L_{j3d} + w_{pj2d}L_{pj2d}.
	\end{aligned}
\end{equation}
	

\begin{table*}\footnotesize
\setlength{\belowcaptionskip}{-25pt}
\setlength{\abovecaptionskip}{-0.5pt}
\renewcommand{\arraystretch}{1.4}
\setlength\tabcolsep{4pt}
\caption{Comparisons to the state-of-the-art methods on 3DPW following \emph{Protocol 1 and 2} (evaluate on the entire 3DPW dataset and on the test set only). * means that additional datasets are used for training~\cite{sun2021monocular} .
}
\label{tab:protocol1-2}
\centerline{\begin{tabular}{ c | c  c  | c | c  c  c}
	\hline
	\multicolumn{3}{c|}{\textbf{\emph{Protocol 1}}} & \multicolumn{4}{c}{\textbf{\emph{Protocol 2}}}\\
	\hline
	
	$\tabincell{c}{\textbf{Methods}}$  & $\tabincell{c}{\textbf{MPJPE}\ \textbf{$\downarrow$} }$ & $\tabincell{c}{\textbf{PAMPJPE}\ \textbf{$\downarrow$}}$ &
	$\tabincell{c}{\textbf{Methods}}$  & $\tabincell{c}{\textbf{MPJPE}\ \textbf{$\downarrow$} }$ & $\tabincell{c}{\textbf{PAMPJPE}\ \textbf{$\downarrow$}}$ &
	$\tabincell{c}{\textbf{PVE}\ \textbf{$\downarrow$}}$\\

    \hline
    
	ROMP(ResNet-50)*~\cite{sun2021monocular} & 87.0 & 62.0 & ROMP(ResNet50)*~\cite{sun2021monocular} & 91.3 & 54.9 & 108.3 \\

	\hline
    \hline
 
	\multirow{2}{*}{Openpose + SPIN~\cite{kolotouros2019learning}} & \multirow{2}{*}{ 95.8 } & \multirow{2}{*}{66.4} 
	 & HMR~\cite{kanazawa2018end} & 130.0 & 76.7 & - \\
	 & & & \cellcolor{gray!10}HMMR~\cite{kanazawa2019learning} & \cellcolor{gray!10}116.5 & \cellcolor{gray!10}72.6 & \cellcolor{gray!10}139.3\\
	  \cellcolor{gray!10}CRMH~\cite{jiang2020coherent} &  \cellcolor{gray!10}105.9 &  \cellcolor{gray!10}71.8 & Arnab \emph{et al.}~\cite{arnab2019exploiting} & - & 72.2 & -\\
 	\multirow{2}{*}{YOLO + VIBE\cite{kocabas2020vibe}*} & \multirow{2}{*}{94.7} &\multirow{2}{*}{66.1} 
	 & \cellcolor{gray!10}GCMR~\cite{kolotouros2019convolutional} &\cellcolor{gray!10} - & \cellcolor{gray!10}70.2 &\cellcolor{gray!10} - \\
	 & & & DSD-SATN~\cite{sun2019human} & - & 69.5 & - \\
	 \cellcolor{gray!10}BMP~\cite{zhang2021body}* &  \cellcolor{gray!10}104.1 &  \cellcolor{gray!10}63.8 & \cellcolor{gray!10}SPIN~\cite{kolotouros2019learning} & \cellcolor{gray!10}96.5 & \cellcolor{gray!10}59.2 & \cellcolor{gray!10}116.4 \\
  
	\hline
 
	\cellcolor{gray!10}ROMP~\cite{sun2021monocular} & \cellcolor{gray!10}90.87 & \cellcolor{gray!10}61.34 & \cellcolor{gray!10}ROMP~\cite{sun2021monocular} & \cellcolor{gray!10}96.96 & \cellcolor{gray!10}57.48 & \cellcolor{gray!10}\textbf{110.13} \\
	CoordFormer(Ours) & \textbf{88.95} & \textbf{59.86} & CoordFormer(Ours) & \textbf{95.27} & \textbf{54.58} & 110.35\\
	\hline
\end{tabular}}
\vspace{-0.5cm}
\end{table*}

\vspace{-4mm}
\section{Experiments}
\vspace{-1mm}
\subsection{Implementation Details}
\vspace{-1mm}

\textbf{Network Architecture.} 
To facilitate a fair comparison, we follow prior approaches~\cite{sun2021monocular} and leverage the HRNet-32~\cite{cheng2020higherhrnet} as the backbone, similar to \cite{sun2021monocular,yuan2022glamr}.

\textbf{Datasets.} To ensure a fair comparison with previous methods, the training is conducted on well-known datasets. The image dataset that is used to train the spatial branch consists of two 3D pose datasets (MPI-INF-3DHP\cite{mehta2017monocular} and MuCo-3DHP~\cite{mehta2017monocular}) and two in-the-wild 2D pose datasets (MPII~\cite{andriluka20142d} and LSP~\cite{johnson2010clustered,johnson2011learning}), while the video dataset consists of the 3DPW~\cite{von2018recovering} and Human3.6M~\cite{ionescu2013human3} datasets.

\textbf{Evaluation.} Evaluation is performed on the 3DPW~\cite{von2018recovering} dataset as the Human3.6M~\cite{ionescu2013human3} and MPI-INF-3DHP\cite{mehta2017monocular} datasets only contain one person per frame and can thus not be used to assess the performance in multi-person scenarios. Therefore, 3DPW~\cite{von2018recovering} is employed as the main benchmark for evaluating the 3D mesh/joint error. Moreover, we follow \cite{sun2021monocular} and divide the 3DPW dataset into three subsets, namely 3DPW-PC, 3DPW-OC and  3DPW-NC. These subsets represent subsets containing person-person occlusion, object occlusion and non-occluded/truncated cases, respectively, and are used to evaluate the performance under different occlusion scenarios.

Following prior approaches~\cite{kocabas2020vibe,wei2022capturing}, the quantitative performance is evaluated by computing the mean per joint position error (MPJPE), the Procrustes-aligned mean per joint position error (PAMPJPE), and the mean Per Vertex Error (PVE) for each frame.

\textbf{Baselines.} We compare CoordFormer to both single-image-based and video-based baseline methods. For single image-based methods, we include HMR~\cite{kanazawa2018end}, SPIN~\cite{kolotouros2019learning}, CRMH~\cite{jiang2020coherent}, EFT~\cite{joo2020exemplar}, BMP~\cite{zhang2021body}, ROMP~\cite{sun2021monocular} and BEV~\cite{sun2022putting}. For video-based methods, we include HMMR~\cite{kanazawa2019learning}, Doersch \emph{et al.}~\cite{doersch2019sim2real}, DSD-SATN~\cite{sun2019human}, VIBE~\cite{kocabas2020vibe}, TCMR~\cite{choi2021beyond}, MEVA~\cite{zheng20213d}, MPS-Net~\cite{wei2022capturing} and MotionBERT~\cite{zhu2022motionbert}. Note that MotionBERT~\cite{zhu2022motionbert} requires additional 2D skeletons motion information as input.

\begin{figure*}
    \setlength{\belowcaptionskip}{-12pt}
	\centerline{\includegraphics[width=0.98\textwidth]{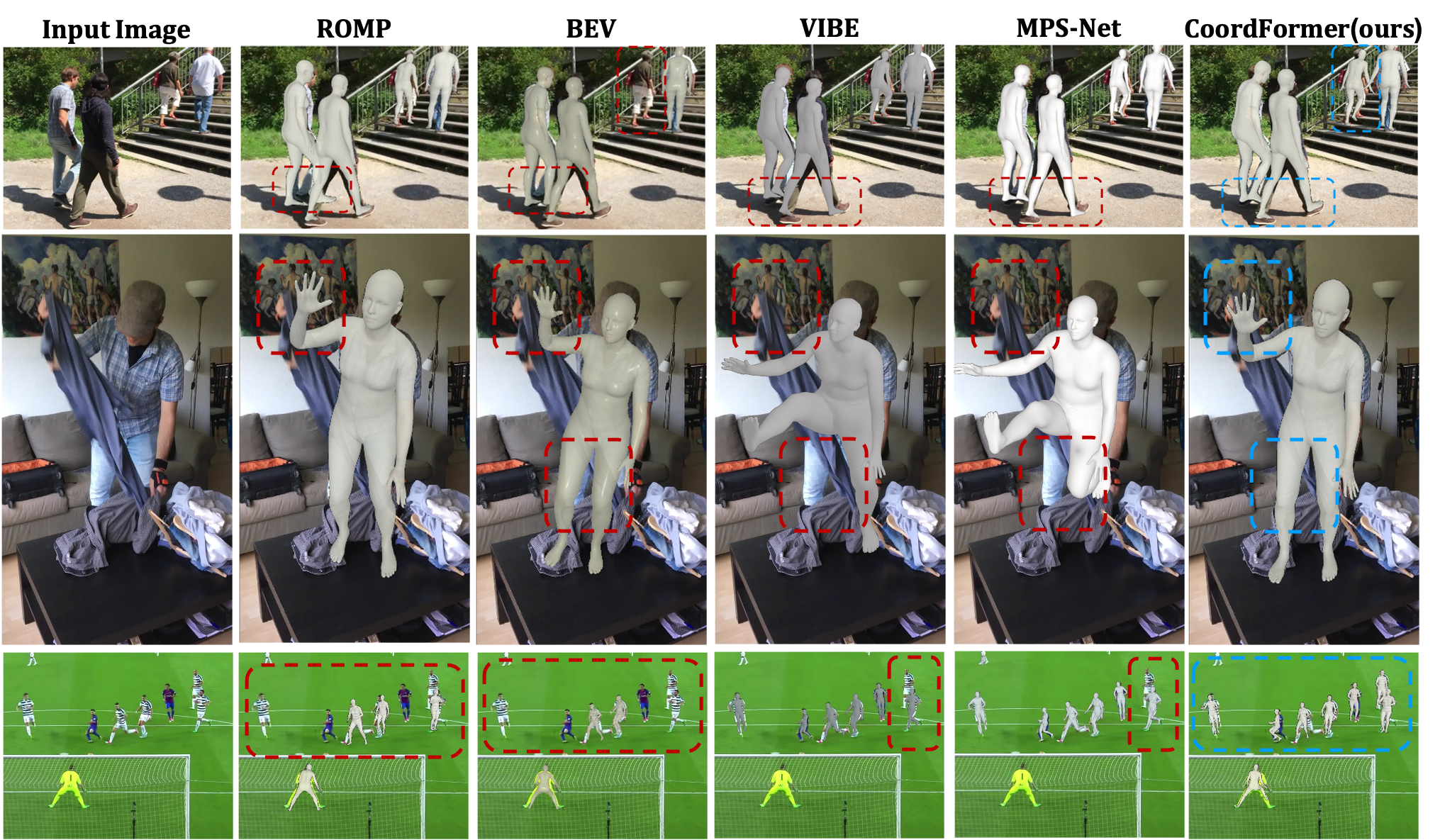}}
	\caption{Qualitative results of ROMP~\cite{sun2021monocular}, BEV~\cite{sun2022putting}, VIBE~\cite{kocabas2020vibe}, MPS-Net~\cite{wei2022capturing} and CordFormer on 3DPW and the internet videos. }
	\label{fig:visual-result}
\end{figure*}

\vspace{-0.3mm}
\subsection{Comparisons to the State-of-the-Art}
    \textbf{In-the-wild multi-person scenarios.} To reveal the effectiveness of CoordFormer, we evaluate CoordFormer under the different in-the-wild scenarios of 3DPW. For a fair and comprehensive comparison, we follow \cite{sun2021monocular} to adopt three evaluation protocols and then compare CoordFormer with state-of-the-art methods. 
    As ROMP was originally trained on a considerably larger dataset, which included OH~\cite{zhang2020object}, the pseduo 3D labels from~\cite{joo2020exemplar}, and PoseTrack~\cite{andriluka2018posetrack}, we retrain ROMP on our dataset to ensure fair comparisons. While we attempted the same with BEV, we observed that BEV did not converge due to the missing relative depth and age supervision that is used to learn BEV's centermap. For completeness, we still report the original results reported in~\cite{sun2021monocular} for both ROMP and BEV as reference.
    
    To comprehensively verify the in-the-wild performance, we follow \emph{Protocol 1} to evaluate models on the entire 3DPW dataset. Without any ground truth as input, single-person methods~\cite{kocabas2020vibe,kolotouros2019learning} are equipped with a 2D human detector~\cite{cao2017realtime,redmon2018yolov3}. As shown in \cref{tab:protocol1-2}, CoordFormer significantly outperforms all the baselines in MPJPE and PAMPJPE, which reveals that CoordFormer can successfully learn the pixel-level feature representation and better model spatial-temporal relations through  ST-Trans.
    
    Moreover, to evaluate the ability in modeling temporal motion constrains, we follow \emph{Protocol 2} on the 3DPW test set without fine-tuning on the 3DPW training set. In \cref{tab:protocol1-2}, CoordFormer takes the whole image as input and the temporal branch is only trained on the Human3.6 M~\cite{ionescu2013human3} dataset, while multi-stage baseline methods can use the cropped single-person image as input and train on more video datasets, i.e. Human3.6 M~\cite{ionescu2013human3}, MPI-INF-3DHP~\cite{mehta2017monocular}, AMASS~\cite{mahmood2019amass}. CoordFormer still outperforms all baselines. Finally, we follow \emph{Protocol 3} to evaluate the models on the 3DPW test set with 3DPW fine-tuning. As shown in \cref{tab:protocol3}, CoordFormer outperforms all the methods in MPJPE and PAMPJPE, while being only slightly worse than MotionBERT in PVE. Note that MotionBERT requires additional 2D skeleton motion as input, while CoordFormer can directly be applied on in-the-wild images.
    
    A qualitative comparisons to state-of-the-art methods is provided in \cref{fig:visual-result}, demonstrating the effectiveness of CoordFormer to precisely recover the mesh. Additional qualitative results are included in the supplementary material. 

    \textbf{3DPW upper-bound performance.}
    To show the upper-bound performance of the video-based methods on the in-the-wild multi-person video dataset, i.e. 3DPW, we compare CoordFormer with previous state-of-the-art video-based methods regardless of their training dataset and training setting. As shown in \cref{tab:comparsion}, CoordFormer achieves the best results, which demonstrates the effectiveness of CoordFormer for multi-person mesh recovery from videos.
    
    \textbf{Occlusion scenarios.} As shown in \cref{tab:occlusion-study}, CoordFormer achieves superior performance on the 3DPW-NC and 3DPW-OC subset under non-occlusion and object occlusion cases according to PAMPJPE. Further comparisons show that CoordFormer outperforms ROMP~\cite{sun2021monocular} in MPJPE on all 3DPW subsets, demonstrating that precise coordinate information improves the performance under occlusion.
    
    \textbf{Runtime comparisons.} In \cref{tab:run-time}, all comparisons are performed on a desktop with a GTX 3090Ti GPU and a Intel(R) Xeon(R) Platinum 8163 CPU. All video-based models are tested on 8-frames video clips. CoordFormer is slightly slower than image-based methods~\cite{sun2021monocular,sun2022putting} due to the overhead in spatial-temporal modeling, however, CoordFormer is significantly faster than the video-based methods \cite{kocabas2020vibe,wei2022capturing}.

\begin{table}\footnotesize
\setlength{\abovecaptionskip}{-1pt}
\setlength{\belowcaptionskip}{-5pt}
\renewcommand{\arraystretch}{1.4}
\setlength\tabcolsep{4pt}
\rowcolors{2}{gray!10}{}
\caption{Comparisons to the state-of-the-art methods on 3DPW following \emph{Protocol 3} (fine-tuned on the training set). * means that additional datasets are used for training~\cite{sun2021monocular}. 
}
\label{tab:protocol3}
\centerline{\begin{tabular}{ c | c  c  c }
	\hline
	$\tabincell{c}{\textbf{Methods}}$  & \tabincell{c}{\textbf{MPJPE}\ \textbf{$\downarrow$} } &
	$\tabincell{c}{\textbf{PAMPJPE}\ \textbf{$\downarrow$}}$ &
	$\tabincell{c}{\textbf{PVE}\ \textbf{$\downarrow$}}$\\

    \hline

	 ROMP(ResNet-50)*~\cite{sun2021monocular} & 84.2 & 51.9 & 100.4\\
	 ROMP(HRNet-32)*~\cite{sun2021monocular} & 78.8 & 48.3 & 94.3\\
	 BEV*~\cite{sun2022putting} & 78.5 & 46.9 & 92.3 \\

	\hline
    \hline
 
	 EFT~\cite{joo2020exemplar} & - & 51.6 & - \\
	 VIBE~\cite{kocabas2020vibe} & 82.9 & 51.9 & 99.1 \\
	 MPS-Net~\cite{wei2022capturing} & 84.3 & 52.1 & 99.7 \\
	 MotionBERT~\cite{zhu2022motionbert} & 80.9 & 49.1 &
	 \textbf{94.2} \\
	
	\hline
 
	ROMP~\cite{sun2021monocular} & 81.06  & 49.07 & 96.74 \\
	CoordFormer(Ours) & \textbf{79.41} & \textbf{46.58} & 94.44 \\
	\hline
\end{tabular}}
\end{table}


\begin{table}\footnotesize
\setlength{\abovecaptionskip}{-1pt}
\setlength{\belowcaptionskip}{-5pt}
\renewcommand{\arraystretch}{1.4}
\setlength\tabcolsep{4pt}
\rowcolors{2}{gray!10}{}
\vspace{-3mm}
\caption{Comparisons of best result to the state-of-the-art video-based methods for in-the-wild scenarios on 3DPW.}
\vspace{1mm}
\label{tab:comparsion}
\centerline{\begin{tabular}{c | c  c  c }
	\hline
$\tabincell{c}{\textbf{Methods}}$ &
	$\tabincell{c}{\textbf{MPJPE}\ \textbf{$\downarrow$} }$ & $\tabincell{c}{\textbf{PAMPJPE}\ \textbf{$\downarrow$}}$ &
	$\tabincell{c}{\textbf{PVE}\ \textbf{$\downarrow$}}$\\
	 
	\hline
	HMMR \cite{kanazawa2019learning} & 116.5 & 72.6 & 139.3 \\
	Doersch \emph{et al.}~\cite{doersch2019sim2real} & - & 74.7 & - \\
	Arnab \emph{et al.}~\cite{arnab2019exploiting} & - & 72.2 & - \\
	DSD-SATN~\cite{sun2019human} & - & 69.5 & - \\
	VIBE \cite{kocabas2020vibe} & 82.9 & 51.9 & 99.1 \\
	MEVA \cite{zheng20213d} & 86.9 & 54.7 & - \\
	TCMR \cite{choi2021beyond} & 86.5 & 52.7 & 103.2 \\
    GLAMR~\cite{yuan2022glamr} + SPEC~\cite{kocabas2021spec} & - & 54.9 & - \\
	GLAMR~\cite{yuan2022glamr} + KAMA~\cite{kaufmann2020convolutional} & - & 51.1 & - \\
	MPS-Net \cite{wei2022capturing} &  84.3 &  52.1 & 99.7\\
	
	CoordFormer(Ours) & \textbf{79.41} & \textbf{46.58} & \textbf{94.44} \\
	
	\hline
\end{tabular}}
\end{table}

\begin{table}\footnotesize
\setlength{\abovecaptionskip}{-1pt}
\setlength{\belowcaptionskip}{-5pt}
\renewcommand{\arraystretch}{1.4}
\setlength\tabcolsep{4pt}
\vspace{-4mm}
\caption{Comparisons to state-of-the-art methods on the person-occluded (3DPW-PC), object-occluded (3DPW-OC) and non-occluded/truncated (3DPW-NC) subsets of 3DPW. * means that additional datasets are used for training.}
\vspace{5pt}
\label{tab:occlusion-study}
\centerline{\begin{tabular}{ c | c  c  c  c }
	\hline
	$\tabincell{c}{\textbf{Metric}}$  &
	$\tabincell{c}{\textbf{Method}}$  &
	$\tabincell{c}{\textbf{3DPW-PC}\ \textbf{$\downarrow$} }$ & $\tabincell{c}{\textbf{3DPW-NC}\ \textbf{$\downarrow$}}$ &
	$\tabincell{c}{\textbf{3DPW-OC}\ \textbf{$\downarrow$}}$ \\
  
	\hline
    \multirow{5}{*}{PAMPJPE} & ROMP* & 75.8 & 57.1 & 67.1\\

    \cline{2-5}

     & CRMH~\cite{jiang2020coherent} & 103.55 & 65.7 &  78.9 \\
	& \cellcolor{gray!10} VIBE~\cite{kocabas2020vibe} & \cellcolor{gray!10} 103.9 & \cellcolor{gray!10} 57.3 & \cellcolor{gray!10} 65.9 \\
     & ROMP  & \textbf{77.64} & 56.67 & 66.6\\
    & \cellcolor{gray!10} CoordFormer & \cellcolor{gray!10} 79.30 & \cellcolor{gray!10} \textbf{54.13} & \cellcolor{gray!10} \textbf{64.47} \\
 
	\hline
    \multirow{2}{*}{MPJPE} & ROMP & 103.70 & 95.53 & 100.79\\
	& \cellcolor{gray!10} CoordFormer & \cellcolor{gray!10} \textbf{101.51} & \cellcolor{gray!10} \textbf{93.17} & \cellcolor{gray!10} \textbf{97.25} \\
	\hline
	
\end{tabular}}
\vspace{-3mm}
\end{table}

\begin{table}\footnotesize
\setlength{\abovecaptionskip}{-1pt}
\setlength{\belowcaptionskip}{-10pt}
\renewcommand{\arraystretch}{1.4}
\setlength\tabcolsep{4pt}
\rowcolors{2}{gray!10}{}
\caption{Run-time comparison on a 3090 GPU.}
\label{tab:run-time}
\centerline{\begin{tabular}{ c | c  c  c  c }
	\hline
	$\tabincell{c}{\textbf{Methods}}$  &
	$\tabincell{c}{\textbf{Time}\\\textbf{per frame(s)}}\downarrow$ & 
	$\tabincell{c}{\textbf{FPS}$\uparrow$}$ &
	$\tabincell{c}{\textbf{Backbbone}}$ &
	$\tabincell{c}{\textbf{Using Temporal}\\\textbf{information}}$ \\
	 
	\hline
	ROMP~\cite{sun2021monocular} & \textbf{0.01329} & \textbf{75.26} & HRNet-32 & \ding{53}\\
	BEV~\cite{sun2022putting} & 0.01448 & 69.04 & HRNet-32& \ding{53}\\
	\hline
	VIBE~\cite{kocabas2020vibe} & 0.07881 & 12.68 & HRNet-32 & \checkmark \\
	MPS-Net~\cite{wei2022capturing} & 0.08013 & 12.47 & HRNet-32 & \checkmark\\
	CoordFormer & \textbf{0.01867} & \textbf{53.55} & HRNet-32 & \checkmark\\
	\hline
	
\end{tabular}}
\vspace{-0.2cm} 
\end{table}

\subsection{Ablation Study}

\begin{table}\footnotesize
\setlength{\abovecaptionskip}{-1pt}
\renewcommand{\arraystretch}{1.4}
\setlength\tabcolsep{4pt}
\rowcolors{2}{gray!10}{}
\caption{Ablation study under 3DPW \emph{Protocol 3}.}
\label{tab:ablation-study-2}
\centerline{\begin{tabular}{ c | c  c  c  c }
	\hline
	$\tabincell{c}{\textbf{Methods}}$  & 
	$\tabincell{c}{\textbf{MPJPE}\ \textbf{$\downarrow$} }$ & $\tabincell{c}{\textbf{PAMPJPE}\ \textbf{$\downarrow$}}$ &
	$\tabincell{c}{\textbf{PVE}\ \textbf{$\downarrow$}}$ \\
	 
	\hline
	CoordFormer w/o CAA & 83.19 & 50.62 & 99.21\\
	CoordFormer w/o BCA & 82.20 & 48.84 & 98.23\\
	CoordFormer & \textbf{79.41} & \textbf{46.58} & \textbf{94.44} \\
	\hline
	
\end{tabular}}
\vspace{-1mm}
\end{table}

\begin{table}\footnotesize
\setlength{\abovecaptionskip}{-1pt}
\setlength{\belowcaptionskip}{-5pt}
\renewcommand{\arraystretch}{1.4}
\setlength\tabcolsep{4pt}
\caption{Ablation study of spatial and temporal Transformer on 3DPW. S means only training the spatial branch, ST means fine-tuning the temporal branch on Human3.6 M, ST-fine means fine-tuning on the 3DPW training set.}
\label{tab:ablation-study-1}
\centerline{\begin{tabular}{ c | c  c  c  c }
	\hline
	$\tabincell{c}{\textbf{Evaluation}}$  & $\tabincell{c}{\textbf{Methods}}$ & 
	$\tabincell{c}{\textbf{MPJPE}\ \textbf{$\downarrow$} }$ & $\tabincell{c}{\textbf{PAMPJPE}\ \textbf{$\downarrow$}}$ &
	$\tabincell{c}{\textbf{PVE}\ \textbf{$\downarrow$}}$ \\
	 
	\hline
	\multirow{2}{*}{On entire 3DPW} & S & 95.05 & 63.22 & 115.90 \\
	& \cellcolor{gray!10} ST & \cellcolor{gray!10} 88.95 & \cellcolor{gray!10} 59.86 & \cellcolor{gray!10} 103.88 \\
	\hline
	
	\multirow{3}{*}{On test set only} & S & 103.95 & 58.03 & 120.67 \\
	& \cellcolor{gray!10} ST & \cellcolor{gray!10} 95.27 & \cellcolor{gray!10} 54.58 & \cellcolor{gray!10} 110.35 \\
	& ST-fine & \textbf{79.41} & \textbf{46.58} & \textbf{94.44} \\
	\hline
	
\end{tabular}}
\vspace{-0.5cm}
\end{table}

To validate the effectiveness of the BCA and CAA modules in CoordFormer, we train CoordFormer under different settings and conduct ablation studies following \emph{Protocol 3} to evaluate on 3DPW. Specifically, we evaluate the BCA module by replacing $\textbf{F}_{focus}$ with $\textbf{F}_{coord}$ without extra attention mechanism and evaluate the CAA module by skipping the Coordinate encoding in \cref{fig:CAA}. As shown in \cref{tab:ablation-study-2}, CoordFormer with BCA and CCA achieves the best result in the in-the-wild scenarios, which fully demonstrates the effectiveness of BCA and CCA. Specifically, the results confirm that BCA can effectively enhance the perception of potential people in the multi-person scenario. Second, the ablation experiments strongly reflect the importance of precise coordinate information in videos. In summary, the results from \cref{tab:ablation-study-2} reveal the importance of capturing position information in the multi-person scenario and the effectiveness of the BCA and CCA modules. We further perform an additional ablation study on the Spatial-Temporal Transformer of CoordFormer. Results in \cref{tab:ablation-study-1} illustrate the benefit of exploiting temporal and spatial information jointly. The reason for the decline in performance when only leveraging the spatial branch can be attributed to two factors: the inability to utilize temporal information and the fact that CAA lacks temporal coordinate information.

\vspace{-3mm}
\section{Conclusion}
\vspace{-2mm}
We proposed CoordFormer to achieve single-stage multi-person mesh recovery from videos. CoordFormer incorporates implicit multi-person detection, tracking, and spatial-temporal modeling. Two critical novelties are the Coordinate-Aware Attention mechanism for pixel-level feature learning and the Body Center Attention for person-focused feature selection. 
CoordFormer paves the way for various downstream applications related to perceiving group behavior, including but not limited to virtual reality and physical therapy.

Despite CoordFormer's robust performance to recover multi-person meshes, its current version lacks the ability to recover completely occluded meshes. We plan to explore this exciting area by leveraging the continuity along the temporal dimension of the body center heatmap.

\vspace{-6mm}
\paragraph{Acknowledgment:} This work was supported in part by National Key R\&D Program of China under Grant No. 2020AAA0109700, Guangdong Outstanding Youth Fund (Grant No. 2021B1515020061), Shenzhen Science and Technology Program (Grant No. RCYX20200714114642083), Shenzhen Fundamental Research Program(Grant No. JCYJ20190807154211365), Nansha Key RD Program under Grant No.2022ZD014 and Sun Yat-sen University under Grant No. 22lgqb38 and 76160-12220011. We thank MindSpore for the partial support of this work, which is a new deep learning computing framwork\footnote{https://www.mindspore.cn/}.

{\small
	\bibliographystyle{ieee_fullname}
	\bibliography{egbib}

\begin{thebibliography}{10}\itemsep=-1pt

\bibitem{andriluka2018posetrack}
Mykhaylo Andriluka, Umar Iqbal, Eldar Insafutdinov, Leonid Pishchulin, Anton
  Milan, Juergen Gall, and Bernt Schiele.
\newblock Posetrack: A benchmark for human pose estimation and tracking.
\newblock In {\em CVPR}, pages 5167--5176. IEEE, 2018.

\bibitem{andriluka20142d}
Mykhaylo Andriluka, Leonid Pishchulin, Peter Gehler, and Bernt Schiele.
\newblock 2d human pose estimation: New benchmark and state of the art
  analysis.
\newblock In {\em CVPR}, pages 3686--3693. IEEE, 2014.

\bibitem{arnab2019exploiting}
Anurag Arnab, Carl Doersch, and Andrew Zisserman.
\newblock Exploiting temporal context for 3d human pose estimation in the wild.
\newblock In {\em CVPR}, pages 3395--3404. IEEE, 2019.

\bibitem{ba2016layer}
Jimmy~Lei Ba, Jamie~Ryan Kiros, and Geoffrey~E Hinton.
\newblock Layer normalization.
\newblock {\em arXiv}, 2016.

\bibitem{bogo2016keep}
Federica Bogo, Angjoo Kanazawa, Christoph Lassner, Peter Gehler, Javier Romero,
  and Michael~J Black.
\newblock Keep it smpl: Automatic estimation of 3d human pose and shape from a
  single image.
\newblock In {\em ECCV}, pages 561--578. Springer, 2016.

\bibitem{cao2017realtime}
Zhe Cao, Tomas Simon, Shih-En Wei, and Yaser Sheikh.
\newblock Realtime multi-person 2d pose estimation using part affinity fields.
\newblock In {\em CVPR}, pages 7291--7299. IEEE, 2017.

\bibitem{cheng2020higherhrnet}
Bowen Cheng, Bin Xiao, Jingdong Wang, Honghui Shi, Thomas~S Huang, and Lei
  Zhang.
\newblock Higherhrnet: Scale-aware representation learning for bottom-up human
  pose estimation.
\newblock In {\em CVPR}, pages 5386--5395. IEEE, 2020.

\bibitem{choi2021beyond}
Hongsuk Choi, Gyeongsik Moon, Ju~Yong Chang, and Kyoung~Mu Lee.
\newblock Beyond static features for temporally consistent 3d human pose and
  shape from a video.
\newblock In {\em CVPR}, pages 1964--1973. IEEE, 2021.

\bibitem{choutas2022accurate}
Vasileios Choutas, Lea M{\"u}ller, Chun-Hao~P Huang, Siyu Tang, Dimitrios
  Tzionas, and Michael~J Black.
\newblock Accurate 3d body shape regression using metric and semantic
  attributes.
\newblock In {\em CVPR}, pages 2718--2728. IEEE, 2022.

\bibitem{doersch2019sim2real}
Carl Doersch and Andrew Zisserman.
\newblock Sim2real transfer learning for 3d human pose estimation: motion to
  the rescue.
\newblock {\em NeurIPS}, 2019.

\bibitem{dosovitskiy2020image}
Alexey Dosovitskiy, Lucas Beyer, Alexander Kolesnikov, Dirk Weissenborn,
  Xiaohua Zhai, Thomas Unterthiner, Mostafa Dehghani, Matthias Minderer, Georg
  Heigold, Sylvain Gelly, et~al.
\newblock An image is worth 16x16 words: Transformers for image recognition at
  scale.
\newblock {\em arXiv}, 2020.

\bibitem{ionescu2013human3}
Catalin Ionescu, Dragos Papava, Vlad Olaru, and Cristian Sminchisescu.
\newblock Human3. 6m: Large scale datasets and predictive methods for 3d human
  sensing in natural environments.
\newblock {\em TPAMI}, pages 1325--1339, 2013.

\bibitem{jiang2020coherent}
Wen Jiang, Nikos Kolotouros, Georgios Pavlakos, Xiaowei Zhou, and Kostas
  Daniilidis.
\newblock Coherent reconstruction of multiple humans from a single image.
\newblock In {\em CVPR}, pages 5579--5588. IEEE, 2020.

\bibitem{johnson2010clustered}
Sam Johnson and Mark Everingham.
\newblock Clustered pose and nonlinear appearance models for human pose
  estimation.
\newblock In {\em BMVC}, page~5. Citeseer, 2010.

\bibitem{johnson2011learning}
Sam Johnson and Mark Everingham.
\newblock Learning effective human pose estimation from inaccurate annotation.
\newblock In {\em CVPR}, pages 1465--1472. IEEE, 2011.

\bibitem{joo2020exemplar}
Hanbyul Joo, Natalia Neverova, and Andrea Vedaldi.
\newblock Exemplar fine-tuning for 3d human pose fitting towards in-the-wild 3d
  human pose estimation.
\newblock In {\em ECCV}, pages 68--84. IEEE, 2020.

\bibitem{kanazawa2018end}
Angjoo Kanazawa, Michael~J Black, David~W Jacobs, and Jitendra Malik.
\newblock End-to-end recovery of human shape and pose.
\newblock In {\em CVPR}, pages 7122--7131. IEEE, 2018.

\bibitem{kanazawa2019learning}
Angjoo Kanazawa, Jason~Y Zhang, Panna Felsen, and Jitendra Malik.
\newblock Learning 3d human dynamics from video.
\newblock In {\em CVPR}, pages 5614--5623. IEEE, 2019.

\bibitem{kaufmann2020convolutional}
Manuel Kaufmann, Emre Aksan, Jie Song, Fabrizio Pece, Remo Ziegler, and Otmar
  Hilliges.
\newblock Convolutional autoencoders for human motion infilling.
\newblock In {\em 3DV}, pages 918--927. IEEE, 2020.

\bibitem{kocabas2020vibe}
Muhammed Kocabas, Nikos Athanasiou, and Michael~J Black.
\newblock Vibe: Video inference for human body pose and shape estimation.
\newblock In {\em CVPR}, pages 5253--5263. IEEE, 2020.

\bibitem{kocabas2021spec}
Muhammed Kocabas, Chun-Hao~P Huang, Joachim Tesch, Lea M{\"u}ller, Otmar
  Hilliges, and Michael~J Black.
\newblock Spec: Seeing people in the wild with an estimated camera.
\newblock In {\em ICCV}, pages 11035--11045. IEEE, 2021.

\bibitem{kolotouros2019learning}
Nikos Kolotouros, Georgios Pavlakos, Michael~J Black, and Kostas Daniilidis.
\newblock Learning to reconstruct 3d human pose and shape via model-fitting in
  the loop.
\newblock In {\em ICCV}, pages 2252--2261. IEEE, 2019.

\bibitem{kolotouros2019convolutional}
Nikos Kolotouros, Georgios Pavlakos, and Kostas Daniilidis.
\newblock Convolutional mesh regression for single-image human shape
  reconstruction.
\newblock In {\em CVPR}, pages 4501--4510. IEEE, 2019.

\bibitem{li2022mhformer}
Wenhao Li, Hong Liu, Hao Tang, Pichao Wang, and Luc Van~Gool.
\newblock Mhformer: Multi-hypothesis transformer for 3d human pose estimation.
\newblock In {\em CVPR}, pages 13147--13156. IEEE, 2022.

\bibitem{liu2018intriguing}
Rosanne Liu, Joel Lehman, Piero Molino, Felipe Petroski~Such, Eric Frank, Alex
  Sergeev, and Jason Yosinski.
\newblock An intriguing failing of convolutional neural networks and the
  coordconv solution.
\newblock {\em NeurIPS}, 2018.

\bibitem{liu2022spatial}
Shuying Liu, Wenbin Wu, Jiaxian Wu, and Yue Lin.
\newblock Spatial-temporal parallel transformer for arm-hand dynamic
  estimation.
\newblock In {\em CVPR}, pages 20523--20532. IEEE, 2022.

\bibitem{loper2015smpl}
Matthew Loper, Naureen Mahmood, Javier Romero, Gerard Pons-Moll, and Michael~J
  Black.
\newblock Smpl: A skinned multi-person linear model.
\newblock {\em TOG}, pages 1--16, 2015.

\bibitem{mahmood2019amass}
Naureen Mahmood, Nima Ghorbani, Nikolaus~F Troje, Gerard Pons-Moll, and
  Michael~J Black.
\newblock Amass: Archive of motion capture as surface shapes.
\newblock In {\em ICCV}, pages 5442--5451. IEEE, 2019.

\bibitem{mehta2017monocular}
Dushyant Mehta, Helge Rhodin, Dan Casas, Pascal Fua, Oleksandr Sotnychenko,
  Weipeng Xu, and Christian Theobalt.
\newblock Monocular 3d human pose estimation in the wild using improved cnn
  supervision.
\newblock In {\em 3DV}, pages 506--516. IEEE, 2017.

\bibitem{pavlakos2019texturepose}
Georgios Pavlakos, Nikos Kolotouros, and Kostas Daniilidis.
\newblock Texturepose: Supervising human mesh estimation with texture
  consistency.
\newblock In {\em CVPR}, pages 803--812. IEEE, 2019.

\bibitem{redmon2018yolov3}
Joseph Redmon and Ali Farhadi.
\newblock Yolov3: An incremental improvement.
\newblock {\em arXiv}, 2018.

\bibitem{sun2021monocular}
Yu Sun, Qian Bao, Wu Liu, Yili Fu, Michael~J Black, and Tao Mei.
\newblock Monocular, one-stage, regression of multiple 3d people.
\newblock In {\em ICCV}, pages 11179--11188. IEEE, 2021.

\bibitem{sun2022putting}
Yu Sun, Wu Liu, Qian Bao, Yili Fu, Tao Mei, and Michael~J Black.
\newblock Putting people in their place: Monocular regression of 3d people in
  depth.
\newblock In {\em CVPR}, pages 13243--13252. IEEE, 2022.

\bibitem{sun2019human}
Yu Sun, Yun Ye, Wu Liu, Wenpeng Gao, Yili Fu, and Tao Mei.
\newblock Human mesh recovery from monocular images via a skeleton-disentangled
  representation.
\newblock In {\em ICCV}, pages 5349--5358. IEEE, 2019.

\bibitem{varol2017learning}
Gul Varol, Javier Romero, Xavier Martin, Naureen Mahmood, Michael~J Black, Ivan
  Laptev, and Cordelia Schmid.
\newblock Learning from synthetic humans.
\newblock In {\em CVPR}, pages 109--117. IEEE, 2017.

\bibitem{vaswani2017attention}
Ashish Vaswani, Noam Shazeer, Niki Parmar, Jakob Uszkoreit, Llion Jones,
  Aidan~N Gomez, {\L}ukasz Kaiser, and Illia Polosukhin.
\newblock Attention is all you need.
\newblock {\em NeurIPS}, 2017.

\bibitem{von2018recovering}
Timo Von~Marcard, Roberto Henschel, Michael~J Black, Bodo Rosenhahn, and Gerard
  Pons-Moll.
\newblock Recovering accurate 3d human pose in the wild using imus and a moving
  camera.
\newblock In {\em ECCV}, pages 601--617. Springer, 2018.

\bibitem{wei2022capturing}
Wen-Li Wei, Jen-Chun Lin, Tyng-Luh Liu, and Hong-Yuan~Mark Liao.
\newblock Capturing humans in motion: Temporal-attentive 3d human pose and
  shape estimation from monocular video.
\newblock In {\em CVPR}, pages 13211--13220. IEEE, 2022.

\bibitem{xu2019denserac}
Yuanlu Xu, Song-Chun Zhu, and Tony Tung.
\newblock Denserac: Joint 3d pose and shape estimation by dense
  render-and-compare.
\newblock In {\em ICCV}, pages 7760--7770. IEEE, 2019.

\bibitem{yuan2022glamr}
Ye Yuan, Umar Iqbal, Pavlo Molchanov, Kris Kitani, and Jan Kautz.
\newblock Glamr: Global occlusion-aware human mesh recovery with dynamic
  cameras.
\newblock In {\em CVPR}, pages 11038--11049. IEEE, 2022.

\bibitem{zeng2022deciwatch}
Ailing Zeng, Xuan Ju, Lei Yang, Ruiyuan Gao, Xizhou Zhu, Bo Dai, and Qiang Xu.
\newblock Deciwatch: A simple baseline for 10x efficient 2d and 3d pose
  estimation.
\newblock {\em arXiv}, 2022.

\bibitem{zhang2021body}
Jianfeng Zhang, Dongdong Yu, Jun~Hao Liew, Xuecheng Nie, and Jiashi Feng.
\newblock Body meshes as points.
\newblock In {\em CVPR}, pages 546--556. IEEE, 2021.

\bibitem{zhang2020object}
Tianshu Zhang, Buzhen Huang, and Yangang Wang.
\newblock Object-occluded human shape and pose estimation from a single color
  image.
\newblock In {\em CVPR}, pages 7376--7385. IEEE, 2020.

\bibitem{zheng20213d}
Ce Zheng, Sijie Zhu, Matias Mendieta, Taojiannan Yang, Chen Chen, and Zhengming
  Ding.
\newblock 3d human pose estimation with spatial and temporal transformers.
\newblock In {\em ICCV}, pages 11656--11665. IEEE, 2021.

\bibitem{zhou2019continuity}
Yi Zhou, Connelly Barnes, Jingwan Lu, Jimei Yang, and Hao Li.
\newblock On the continuity of rotation representations in neural networks.
\newblock In {\em CVPR}, pages 5745--5753. IEEE, 2019.

\bibitem{zhu2022motionbert}
Wentao Zhu, Xiaoxuan Ma, Zhaoyang Liu, Libin Liu, Wayne Wu, and Yizhou Wang.
\newblock Motionbert: Unified pretraining for human motion analysis.
\newblock {\em arXiv}, 2022.

\end{thebibliography}
	
}

\clearpage


\section*{Supplementary material}

# Coordinate Transformer: Achieving Single-stage Multi-person Mesh Recovery from Videos — Supplementary Materials

## 1. Introduction

Here we provide more implementation details of the training and evaluation; additional evaluation results on in-the-wild datasets; additional ablation studies with visualizations; and additional visualization results on internet videos. Further, we perform a comparisons to state-of-the-art methods on extreme scenarios to complement our comprehensive analysis. Finally, a video demo is provided in our supplementary material to show additional qualitative video results.

## 2. Implementation Details

Similar to [12], we resize the image sequences to a size of 512×512. The size of the backbone features is $H_f = W_f = 128$, and the maximum number of detections $N$ is set to 64. The time window size $T$ is set to 8. The spatial loss weights are set to $w_{cm}$ = 160, $w_{mpj}$ = 360, $w_{pmpj}$ = 400, $w_{pj2d}$ = 420, $w_{pose}$ = 80, $w_{shape}$ = 1, $w_{prior}$ = 1.6, and are fixed in all training steps. The temporal loss weights are set to $w_{accel}$ = 200, $w_{aj3d}$ = 300, $w_{sm}$ = 100, and are only used when fine-tuning on the video datasets. The threshold $t_c$ of the Body Center Heatmap is set to 0.2. The learning rate $lr$ of the Adam optimizer is set to 5× $10^{-5}$ and the batch size $B$ is 16. Training is performed in an end-to-end manner directly from image or video inputs.

### 2.1. Training Datasets

Since CoordFormer is a novel approach for multi-person videos, our training and evaluation focuses on the most relevant dataset (3DPW [14]), while other datasets are used as supplements to improve generalization and enhance prediction accuracy. For completeness, we have included the dataset details below.
**3DPW** [14] is a challenging outdoor dataset with more than 51,000 frames of 7 actors in various clothing styles. The dataset includes numerous frames with multi-person interactions and all the raw ground-truth markers are recorded via Inertial Measurement Units, which provide accurate ground truth annotations.
**Human3.6M** [2] is an indoor, multi-view, single-person 3D human pose estimation dataset. The extended SMPL model annotations are generated from sparse marker data. Following [12], we use 5 subjects (S1,S5,S6,S7,S8) for training.
**MPI-INF-3DHP** [10] is an indoor, multi-view, single-person 3D human pose dataset with some noise.
**MuCo-3DHP** [10] is an extended version of MPI-INF-3DHP using data augmentation. The authors replace the background with real-world images and place 1 to 4 subjects on the background to facilitate a range of inter-person overlap and activity scenarios.
**In-the-wild 2D datasets** MPII [1] and LSP [3, 4] are in-the-wild 2D datasets, which are collected using Amazon Mechanical Turk. Annotation quality of the 2D labels is improved by modelling the annotator error using iterative procedures.

### 2.2. Training Steps

Existing video-based methods use an explicit 2D detector and a tracker to model the temporal relationship of a particular individual. Instead, CoordFormer employs Body Center Attention as an implicit detector and the Spatial-Temporal Transformer to learn temporal relations. This allows CoordFormer to not only leverage video data, which often can be restricted in the multi-person setting, but also leverage available image datasets. This can be facilitated by training CoordFormer in three steps: First the spatial branch of CoordFormer will be trained like most existing single image-based methods [12, 8], while the second step consists of fine-tuning the spatial and temporal branch on the video dataset without 3DPW. Finally, CoordFormer is fine-tuned with the 3DPW dataset in the third step.

### 2.3. Training Dataset Ratio

To obtain the best results, we follow EFT [6] and SPIN [9] to batch data according to the dataset sample ratios. To train the spatial branch of CoordFormer without using temporal information, we incorporate 30% MPI-INF-3DHP, 10% LSP, 15% MPII, 20% MuCo-3DHP and 25% Human3.6 M into training in the first step. Next, we fine-tune the model on the Human3.6 M dataset in the second step. Finally, the model is fine-tuned on 3DPW to achieve the final best performing model.

## 2.4. Evaluation Strategy

For a fair comparison, we use CoordFormer after the second training-step for evaluation following *Protocol 1 and 2*, and use the fine-tuned model after the third training-step to evaluate the best results (*Protocol 3*).

Table 1: Comparsions resuls on CMU Panoptic[5] and MuPoTs[11] according to PAMPJPE metric.

| Methods | CMU Panoptic | | | | | MuPoTs |
| --- | --- | --- | --- | --- | --- | --- |
| | Haggling | Mafia | Ultim | Pizza | Mean | |
| ROMP | 68.16 | 79.25 | **76.89** | 85.25 | 77.39 | 93.00 |
| CoordFormer | **66.82** | **77.23** | 77.83 | **83.03** | **76.23** | **88.02** |

## 3. Further evaluation results on in-the-wild datasets.

According to the PAMPJPE metric, the comparison results on CMU Panoptic[5] and MuPoTs[11] are reported for comprehensive evaluations under in-the-wild multi-person scenarios. All the methods are directly evaluated without any fine-tuning. As shown in Tab. 1, CoordFormer outperforms ROMP in almost all activities. Moreover, as shown in Fig. 1, CoordFormer performs better detection and pose estimation.

## 4. Further Ablation Study With Visualization

To further show the effectiveness of BCA and CAA, we collect the evaluation results on the 3DPW validation set during the training and compare the performance of the ablation models. All the models are trained for 10 epochs, following the same setting as the first training step in Sec. 2.2. For the sake of readability, the notations of different model settings are summarized in Tab. 2.

**How BCA accelerates the training process.** As shown in Fig. 2b and Fig. 2c, models with BCA mechanism achieve better performance and more stable convergence under different model settings. More specially, $CF_{bca}$ outperforms $CF_{None}$ by 8.2% and 6.8% according to the MPJPE and PAMPJPE metrics, while $S\text{-}CF_{None}$ could not converge for the same training datasets.

Table 2: The notations of CoordFormer under different settings. Note, here "splitting" refers to adopting the tokenization method that splits the features into patches and extract tokens from them and CAA is not used to highlight the effectiveness of BCA.

| | w/o BCA | w/ BCA |
| --- | --- | --- |
| splitting | $S\text{-}CF_{None}$ | $S\text{-}CF_{bca}$ |
| not splitting | $CF_{None}$ | $CF_{bca}$ |

Table 3: Ablation study of CAA under different training steps.

| Steps | Methods | MPJPE↓ | PAMPJPE↓ | PVE↓ |
| --- | --- | --- | --- | --- |
| Step 1 | $CF_{bca}$ | **101.73** | **55.68** | **117.91** |
| | CoordFormer | 103.95 | 58.03 | 120.67 |
| Step 2 | $CF_{bca}$ | 97.14 | 56.01 | 112.69 |
| | CoordFormer | **95.27** | **54.58** | **110.35** |
| Step 3 | $CF_{bca}$ | 83.19 | 50.62 | 99.21 |
| | CoordFormer | **79.41** | **46.58** | **94.44** |

**How CAA preserves pixel-level representations.** Patch-level tokenization of standard vision transformers leads to feature disorder and feature partition, which occurs when one patch contains multiple person and when the body center is located on the boundary line of the patch, respectively. The design of CAA allow us to keep pixel-level representations and avoid the spatial information degradation. As shown in Fig. 2d, $S\text{-}CF_{None}$ results in extensive fluctuations in performance and crashes halfway through training due to sudden excessive losses. Moreover, we observe that $CF_{None}$ converges quicker than $S\text{-}CF_{bca}$.

However, it is not enough to build spatial-temporal constraints only based on pixel-level tokenization. Tab. 3 illustrates CAA's effectiveness to capture coordinate information across frames. Although $CF_{bca}$ performs better for single-image regression, CoordFormer demonstrates superior modeling of spatial-temporal relations.

**Qualitative ablation study of visualization comparison.** Fig. 3 illustrates that 1) BCA and CAA have to be combined to facilitate accurate Body Center heatmap prediction using the CoordFormer, 2) BCA can enhance the confidence on the body center, especially under person-occlusion scenarios, 3) CoordFormer w/o CAA regresses the mesh only based on BCA, which improves results on certain individuals, but fails to model temporal relations, thus degrading pose and shape coherence.

**Whether BCA and CAA conform to our assumptions.** As shown in Tab. 4, CoordFormer with CAA improves performance by learning temporal information in training step 2 according to both MPJPE and PAMPJPE, while CoordFormer with only BCA obtains worse PAMPJPE. This illustrates that BCA focuses on single-image and CAA focuses on temporal information, which is consistent with our implicit detection and tracking assumptions. Moreover, as shown in Fig. 1, CoordFormer performs better detection and pose estimation.

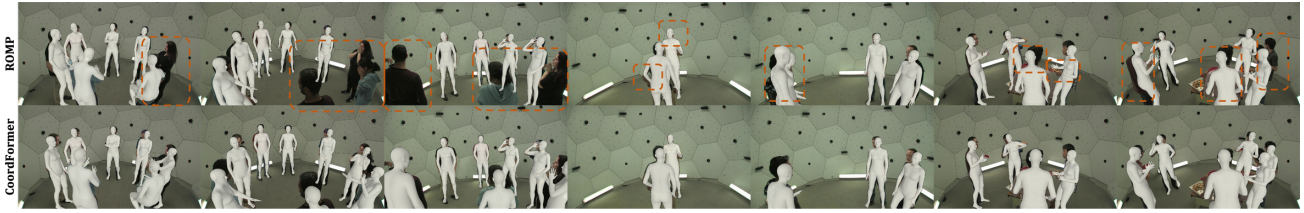

Figure 1: Partial visualization comparison between CoordFormer and ROMP on CMU Panoptic[5] dataset.

Table 4: Ablation study under 3DPW.

| Methods | | MPJPE ↓ | | | PAMPJPE ↓ | | | PVE ↓ | | |
|---|---|---|---|---|---|---|---|---|---|---|
| BCA | CAA | step1 | step2 | step3 | step1 | step2 | step3 | step1 | step2 | step3 |
| ✓ | | **101.73** | 97.14 | 83.19 | **55.68** | 56.01 | 50.62 | **117.91** | 112.69 | 99.20 |
| | ✓ | 103.61 | 99.94 | 82.20 | 57.64 | 55.63 | 48.84 | 120.04 | 114.82 | 98.23 |
| ✓ | ✓ | 103.95 | **95.27** | **79.41** | 58.03 | **54.58** | **46.58** | 120.67 | **110.35** | **94.44** |

## 5. Further visualization results on the internet videos.

We further test CoordFormer on internet videos, especially sports videos. As shown in Fig. 4, CoordFormer effectively obtains the multi-person mesh from a variety of videos.

## 6. Further Visualization Compared to State-of-the-art Methods.

To show the superior performance of CoordFormer beyond simple in-the-wild scenarios, we compare CoordFormer to the best pre-trained ROMP [12] and BEV [13] models. Note, these were trained on a considerably larger number of datasets and for BEV, leverage a larger Body Center heatmap with a size of 128, resulting in more capacity to provide accurate and precise predictions. While an unfair comparison from CoordFormer's perspective, we observe that CoordFormer still obtains preferable results.

**Qualitative results on internet videos with small targets.** Given the precise coordinate information to refine the Body Center heatmap, CoordFormer is able to better detect people in the video, especially the small targets, which is crucial for 3D human mesh recovery from athletic sports videos and aerial videos. As shown in Fig. 5, CoordFormer obtains great detection results and achieves the best visualization results for small targets. While CoordFormer is not able to provide mesh results for all people, CoordFormer achieves significant improvements on the accuracy of the Body Center heatmap and Camera map compared to ROMP and BEV.

**Qualitative results on internet videos with low resolution.** As shown in Fig. 6, CoordFormer displays superior robustness to videos with different resolution. Specifically, CoordFormer can still maintain its performance even for videos with low resolution of 64×36. Compared with state-of-the-art video-based [7, 15] methods that are equipped with 2D detectors, Fig. 7 illustrates that CoordFormer achieves the best results over methods with requiring explicit detection.

## 7. Social Impact

While there are a wide range of application domains where video-based 3D human mesh recovery will be beneficial, such as for instance in physical therapy and virtual reality, there are also potentially negative application scenarios. For instance, these approaches could be used in malicious contexts to obtain a large amount of private body data or for surveillance purposes. Consequently, CoordFormer is released as a research tool only.

## References

[1] Mykhaylo Andriluka, Leonid Pishchulin, Peter Gehler, and Bernt Schiele. 2d human pose estimation: New benchmark and state of the art analysis. In *CVPR*, pages 3686–3693. IEEE, 2014. 1

[2] Catalin Ionescu, Dragos Papava, Vlad Olaru, and Cristian Sminchisescu. Human3. 6m: Large scale datasets and predictive methods for 3d human sensing in natural environments. *TPAMI*, pages 1325–1339, 2013. 1

[3] Sam Johnson and Mark Everingham. Clustered pose and nonlinear appearance models for human pose estimation. In *BMVC*, page 5. Citeseer, 2010. 1

[4] Sam Johnson and Mark Everingham. Learning effective human pose estimation from inaccurate annotation. In *CVPR*, pages 1465–1472. IEEE, 2011. 1

[5] Hanbyul Joo, Hao Liu, Lei Tan, Lin Gui, Bart Nabbe, Iain Matthews, Takeo Kanade, Shohei Nobuhara, and Yaser Sheikh. Panoptic studio: A massively multiview system for social motion capture. In *ICCV*, pages 3334–3342, 2015. 2, 3

3

[6] Hanbyul Joo, Natalia Neverova, and Andrea Vedaldi. Exemplar fine-tuning for 3d human pose fitting towards in-the-wild 3d human pose estimation. In *ECCV*, pages 68–84. IEEE, 2020. 1

[7] Muhammed Kocabas, Nikos Athanasiou, and Michael J Black. Vibe: Video inference for human body pose and shape estimation. In *CVPR*, pages 5253–5263. IEEE, 2020. 3, 9

[8] Muhammed Kocabas, Chun-Hao P Huang, Otmar Hilliges, and Michael J Black. Pare: Part attention regressor for 3d human body estimation. In *ICCV*, pages 11127–11137. IEEE, 2021. 1

[9] Nikos Kolotouros, Georgios Pavlakos, Michael J Black, and Kostas Daniilidis. Learning to reconstruct 3d human pose and shape via model-fitting in the loop. In *ICCV*, pages 2252–2261. IEEE, 2019. 1

[10] Dushyant Mehta, Helge Rhodin, Dan Casas, Pascal Fua, Oleksandr Sotnychenko, Weipeng Xu, and Christian Theobalt. Monocular 3d human pose estimation in the wild using improved cnn supervision. In *3DV*, pages 506–516. IEEE, 2017. 1

[11] Dushyant Mehta, Oleksandr Sotnychenko, Franziska Mueller, Weipeng Xu, Srinath Sridhar, Gerard Pons-Moll, and Christian Theobalt. Single-shot multi-person 3d pose estimation from monocular rgb. In *3DV*, pages 120–130. IEEE, 2018. 2

[12] Yu Sun, Qian Bao, Wu Liu, Yili Fu, Michael J Black, and Tao Mei. Monocular, one-stage, regression of multiple 3d people. In *ICCV*, pages 11179–11188. IEEE, 2021. 1, 3, 7, 8, 9

[13] Yu Sun, Wu Liu, Qian Bao, Yili Fu, Tao Mei, and Michael J Black. Putting people in their place: Monocular regression of 3d people in depth. In *CVPR*, pages 13243–13252. IEEE, 2022. 3, 7, 8, 9

[14] Timo Von Marcard, Roberto Henschel, Michael J Black, Bodo Rosenhahn, and Gerard Pons-Moll. Recovering accurate 3d human pose in the wild using imus and a moving camera. In *ECCV*, pages 601–617. Springer, 2018. 1

[15] Wen-Li Wei, Jen-Chun Lin, Tyng-Luh Liu, and Hong-Yuan Mark Liao. Capturing humans in motion: Temporal-attentive 3d human pose and shape estimation from monocular video. In *CVPR*, pages 13211–13220. IEEE, 2022. 3, 9

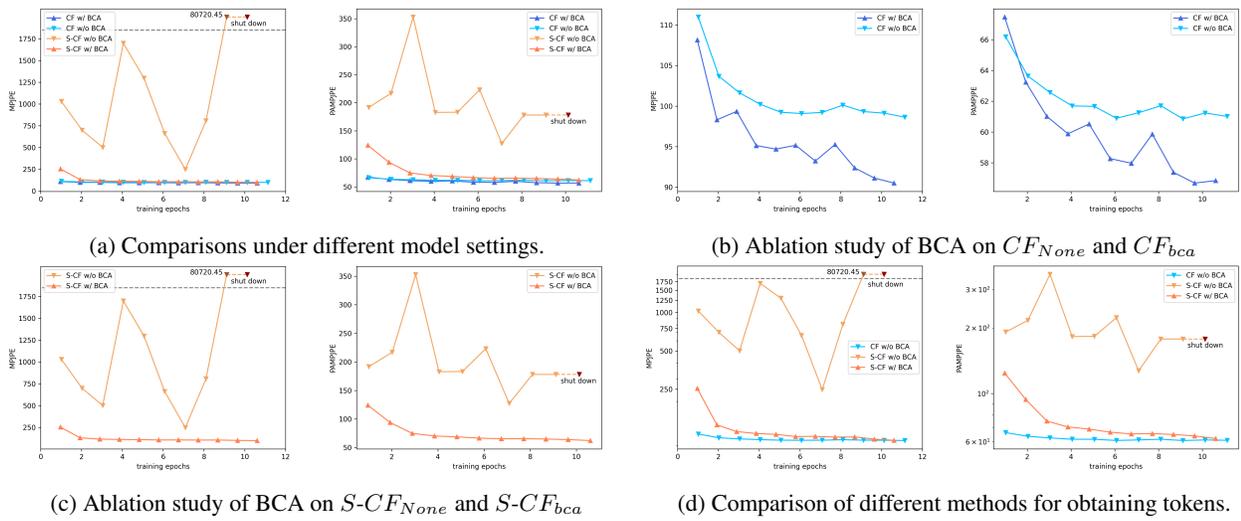

(a) Comparisons under different model settings.

(b) Ablation study of BCA on $CF_{None}$ and $CF_{bca}$

(c) Ablation study of BCA on $S\text{-}CF_{None}$ and $S\text{-}CF_{bca}$

(d) Comparison of different methods for obtaining tokens.

Figure 2: Further ablation study of BCA and CAA at the first training step.

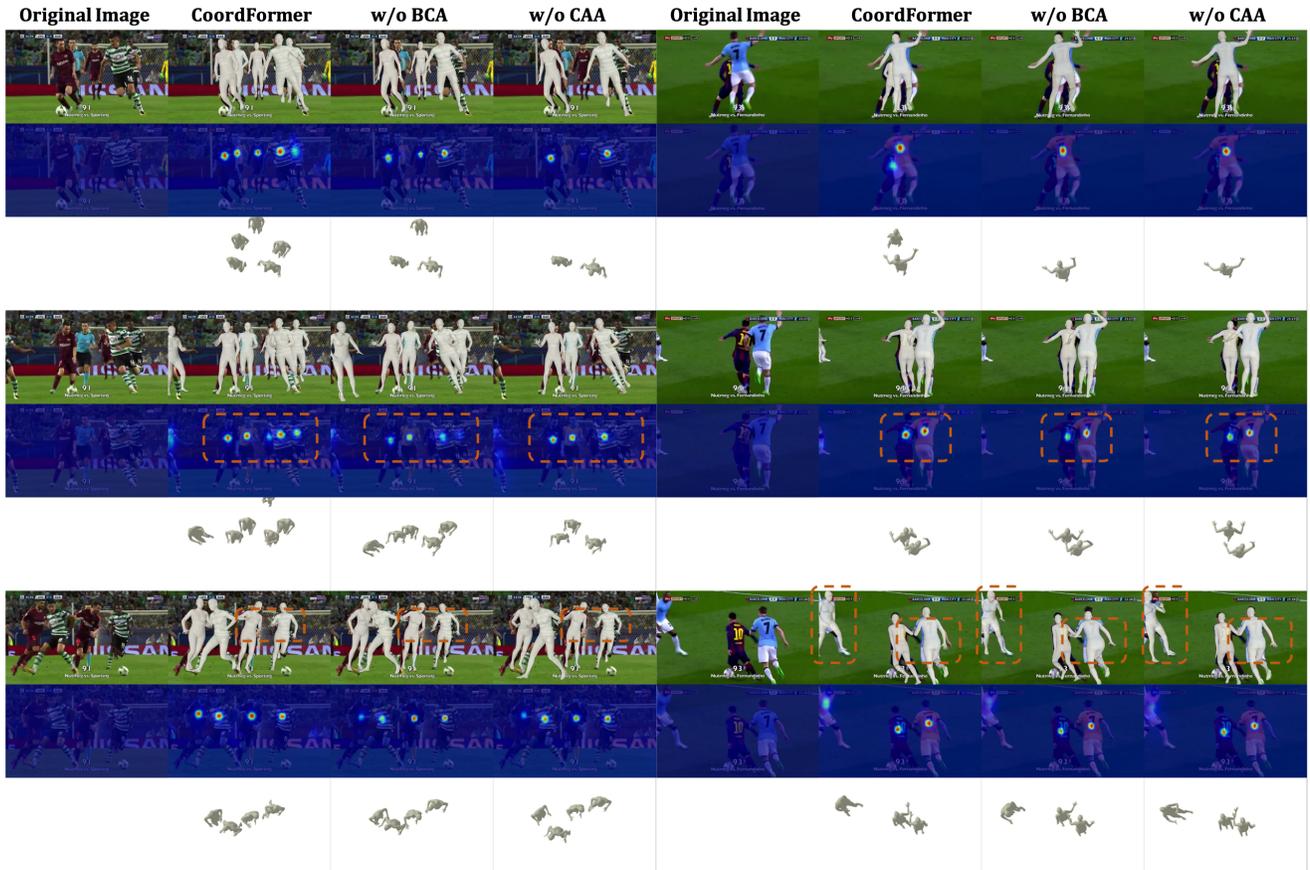

Figure 3: Further ablation study of visualization comparison.

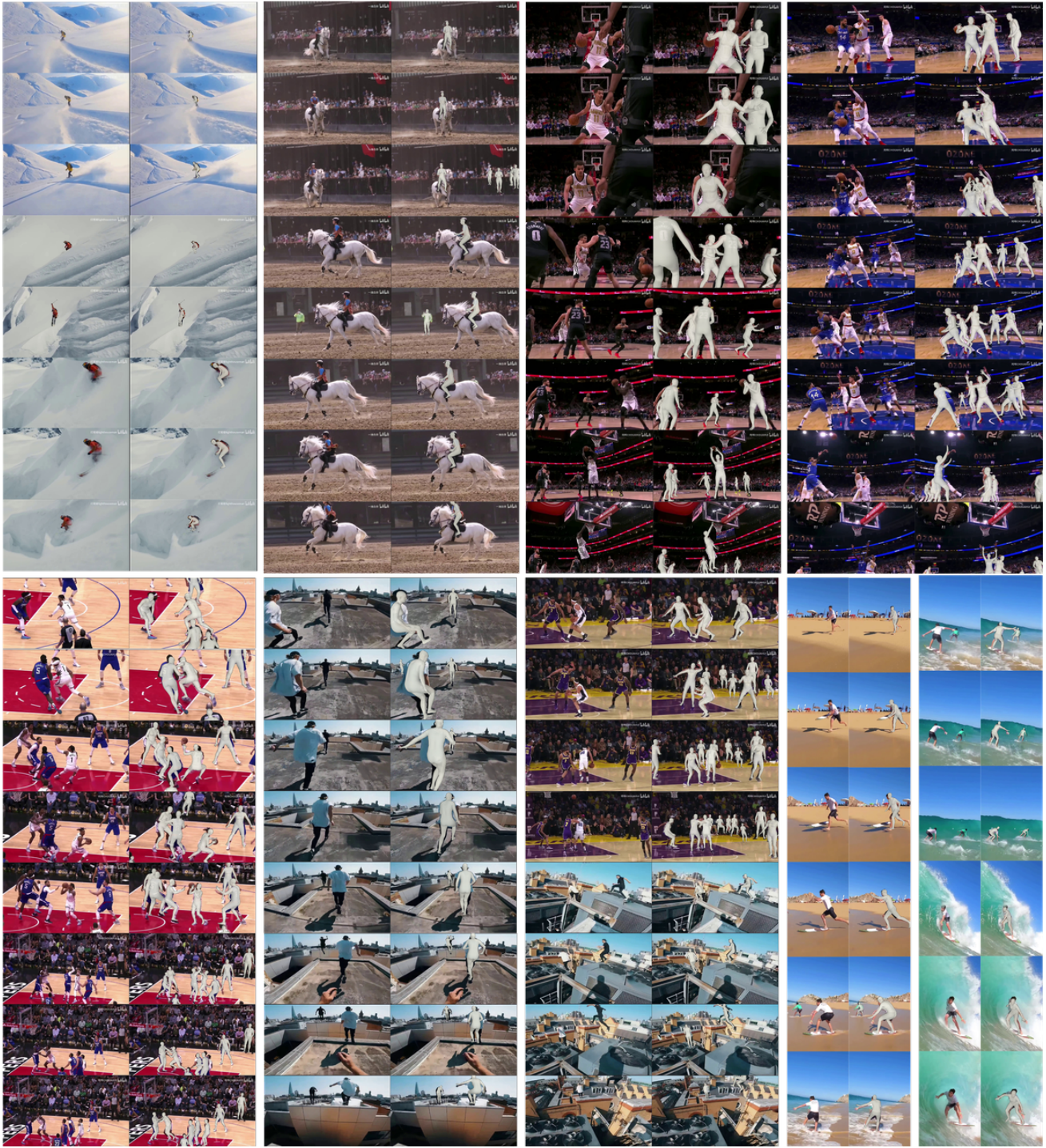

Figure 4: Further visualization results of CoordFormer on the internet videos.

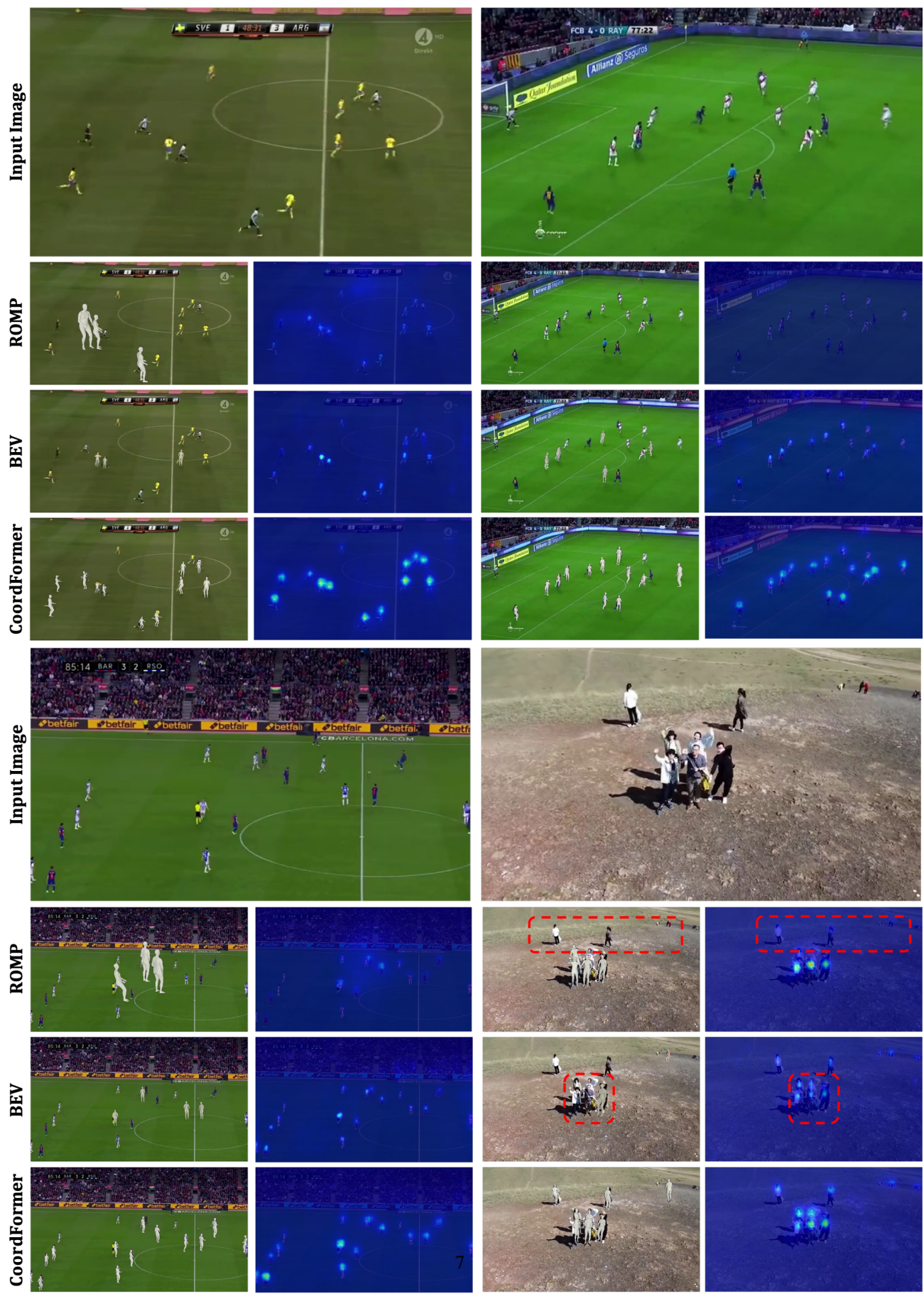

Figure 5: Qualitative results of ROMP [12], BEV [13] and CoordFormer on the internet videos with small targets.

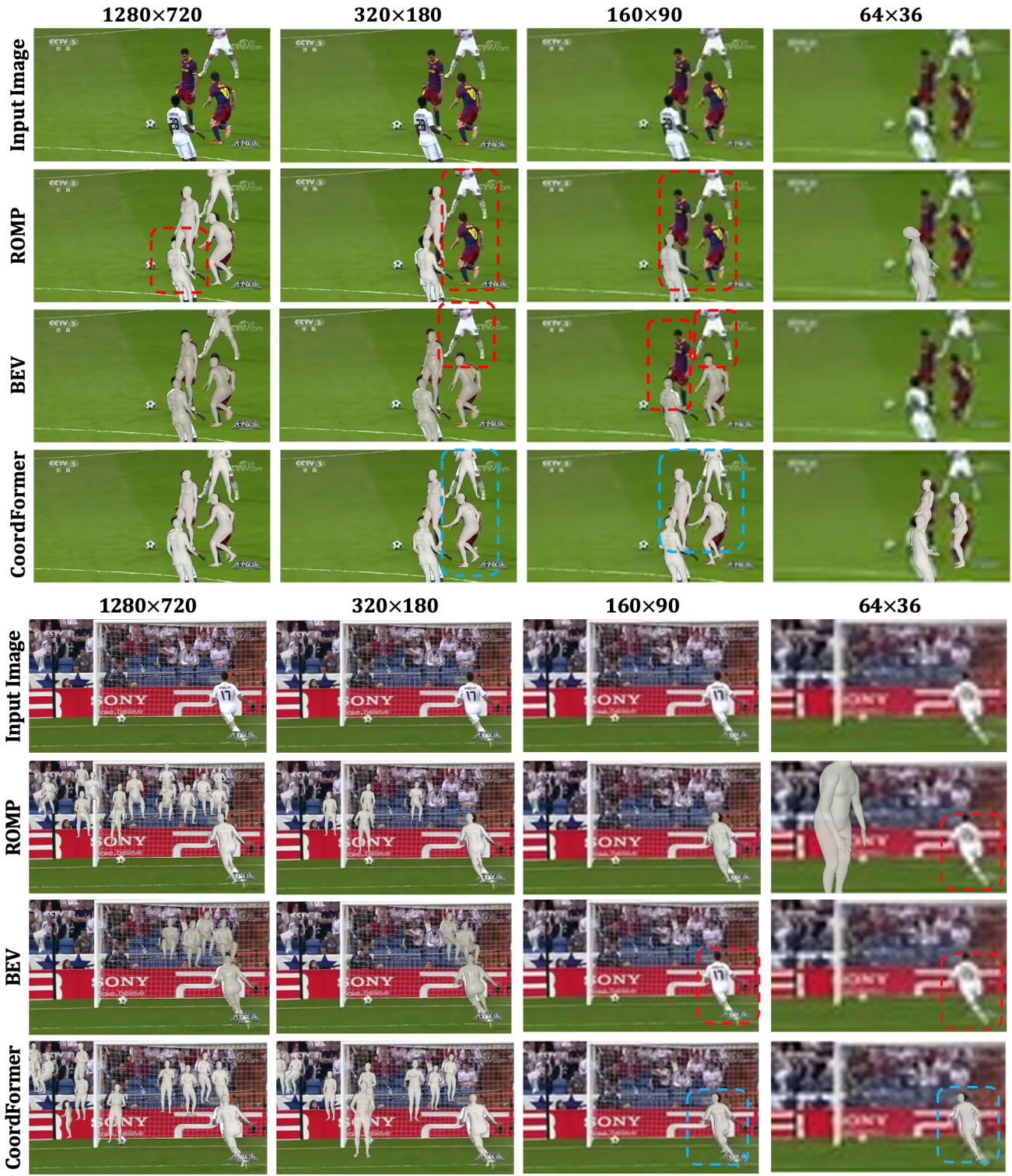

Figure 6: Qualitative results of ROMP [12], BEV [13] and CoordFormer on the internet videos with low resolution.



\end{document}